\documentclass[a4paper,fleqn]{cas-dc}

\usepackage[authoryear,longnamesfirst]{natbib}
\usepackage{amsmath}
\usepackage{booktabs}
\usepackage{graphicx}
\usepackage{subcaption}
\usepackage{xcolor}
\usepackage{amssymb}

\begin{document}
\let\WriteBookmarks\relax
\def\floatpagepagefraction{1}
\def\textpagefraction{.001}

\shorttitle{NL2Hull: Constrained ship design}
\shortauthors{Huo et~al.}

\title [mode = title]{NL2Hull: A Natural Language-Driven Constrained Ship Design Decision Framework}

\author[1]{Wenhua Huo}
\author[1]{Fenglei Han}
\author[1]{Wangyuan Zhao}
\cormark[1]
\ead{heu_wangyuanzhao@163.com}
\author[1]{Jialin Wu}
\author[1]{Jiayi Han}
\affiliation[1]{organization={Harbin Engineering University},
                addressline={145 Nantong Street, Nangang District},
                city={Harbin},
                country={P. R. China}}
\cortext[1]{Corresponding author}

\begin{abstract}
Ship-form design combines smooth geometric representation, local shape editing,
and constraints on the resulting hull. We present the Natural-Language-to-Hull
Framework (NL2Hull Framework), which formulates ship-form editing as a typed
discrete decision problem and connects language decisions to numerical geometry.
Its Constrained Free-Form Deformation Engine (CFFD Engine) represents hull
waterlines with non-uniform rational B-splines (NURBS), applies free-form
deformation (FFD) to their control points, reconstructs the hull, and checks
geometric constraints. We construct the Ship Design Decision Dataset (SDD
Dataset) with 134,558 cleaned records and evaluate compared models on its subset
Ship Design Decision Benchmark (SDDBench), containing 5,000 records and 43,496
typed questions. We propose Chip, a constrained ship-design decision model for processing natural-language requests. Chip reaches 95.90\%
question accuracy and 99.32\% FFD exact match, with a negative log-likelihood of
0.0951, an expected calibration error of 0.0032, and a Brier score of 0.0551.
The NL2Hull Framework provides a reproducible interface for evaluating language-based ship-form decisions while
identifying the geometry and continuous-control components that require further
development.
Our code and dataset is available at \url{https://github.com/wenhuahuo/NL2Hull}.
\end{abstract}

\begin{highlights}
\item Defines a NURBS-based discrete decision problem for constrained ship-form editing.
\item Connects typed probability prediction to executable free-form deformation actions.
\item Provides common benchmark, calibration, action-level, and end-to-end evaluations.
\end{highlights}


\maketitle

\section{Introduction}
\label{sec:introduction}

Ship-form design relies on geometric representations that preserve smoothness
while exposing parameters that naval architects can interpret and modify. Early
parametric hull-form work combined naval-architecture features with non-uniform
rational B-spline (NURBS) surfaces and used control-point manipulation to vary
local hull zones~\citep{nam2000parametric}. Parametric hull-form design also
connected ship-design practice with NURBS skinning and feature deformation
~\citep{abt2001parametric,shamsuddin2006skinning,zhou2022nurbs}.
Parametric ship-design studies further linked hull-form variables to practical
design workflows and objective-driven optimization~\citep{wilson2010hull,peri2001cfd}.
Uncertainty-aware and numerical optimization methods extended this design line
~\citep{campana2009optimization,diez2012uncertainty}.
These studies establish a useful geometric foundation, while also showing that
the quality of an editable hull depends on the choice of parameterization,
continuity across regions, and the relation between geometric parameters and
engineering requirements. A numerical hull representation alone does not specify
which edit a designer intends or how a sequence of edits should be verified.

Free-form deformation (FFD) provides a general mechanism for applying local and
global changes to geometric models, with continuity control and volume-preserving
constructions available within the deformation formulation~\citep{sederberg1986ffd}.
In ship design, parametric generation and deformation have supported hull
optimization and constrained hull generation. ShipGen produces parametric hull
vectors under multiple objectives and constraints~\citep{bagazinski2023shipgen},
and C-ShipGen conditions hull generation on design requirements and resistance
information~\citep{bagazinski2024cshipgen}. These methods focus on generating or
optimizing hull candidates from structured parameters and performance objectives,
with hydrodynamic optimization providing established links between geometry and
performance objectives~\citep{peri2001cfd,campana2009optimization}.
An interactive design system also needs to map a designer's language to a
sequence of localized operations and then verify the resulting geometry after
each sequence.

Natural-language interfaces provide a route to this interaction. Early work on
natural-language computer-aided design established the goal of translating design
intent into parametric operations~\citep{samad1985cad}. Work on valid parametric
CAD models and relational sketch constraints emphasizes that executable design
representations require structural consistency~\citep{hoffmann2001valid,seff2020sketchgraphs}.
DeepCAD learns sequential CAD commands~\citep{wu2021deepcad}, Text2CAD generates
parametric sequences from text~\citep{khan2024text2cad}, and BrepGen models
structured boundary-representation geometry~\citep{xu2024brepgen}.
Visual-feedback training further connects generated sequences to rendered
outcomes~\citep{wang2025cadfusion}.
Language-grounded action systems connect instructions to executable skills and
policy programs~\citep{ahn2022saycan,liang2022codepolicies}. Multimodal embodied
models incorporate observations into planning or action prediction~\citep{driess2023palme,brohan2023rt2},
while closed-loop language feedback supports replanning during execution~\citep{huang2022innermonologue}.
Typed decision models provide another relevant interface: they return choices,
rubric scores, and probabilities for explicitly specified questions instead of
unrestricted text~\citep{deusser2026jev}. Structured-output methods constrain machine-readable output structure during
generation through parsing, grammars, and programmatic constraints
~\citep{scholak2021picard,geng2023grammar,willard2023guided,beurer2022lmql}.
Together, these directions motivate a ship-specific interface in which language
understanding, typed action selection, and geometric execution are evaluated as
linked stages.

To connect design intent with executable ship-form edits, we propose the
Natural-Language-to-Hull Framework (NL2Hull Framework). At its core is Chip, a ship-design decision model for
natural-language editing requests. Chip serves as the decision layer between the designer's
language and the geometric operations. This role is motivated by the structure
of ship-form editing: designers often express changes qualitatively, while each
edit requires a choice of region, operation, magnitude level, and preservation
constraints. We therefore formulate language interpretation as a set of typed
decisions over a predefined action vocabulary. Given a request and these
questions, Chip predicts probability distributions for the action count and
attributes of each ordered action, translating design intent into discrete
editing decisions while retaining uncertainty for calibration assessment.
The highest-probability answers are then converted into actions for the
Constrained Free-Form Deformation Engine (CFFD Engine), which applies them to a
separately supplied NURBS hull representation. Finally, hull reconstruction and
geometric constraint checks connect the predicted decisions to the resulting
ship form. Together, Chip and the CFFD Engine link language interpretation,
sequential editing, and geometric verification within a single framework.

The study makes four contributions:
\begin{enumerate}
    \item We define a NURBS-based decision task for translating
    natural-language ship-form intent into interpretable FFD actions.
    \item We propose Chip, a decision model for ship-form design. It converts
    natural-language requests into ordered discrete decisions and predicts
    probabilities for action choices, qualitative magnitude levels, and
    preservation constraints.
    \item We implement NL2Hull Framework with the CFFD Engine. It connects
    Chip's decisions to sequential deformation, hull reconstruction, and
    geometric constraint checks.
    \item We construct the Ship Design Decision Dataset (SDD Dataset) and its
    Ship Design Decision Benchmark (SDDBench), with a common evaluation protocol
    for question accuracy, probability quality, FFD exact match, and latency
    across Chip, JEV, and general language models. We further validate end-to-end
    editing with Chip.
\end{enumerate}

\begin{figure*}[t]
    \centering
    \includegraphics[width=\textwidth]{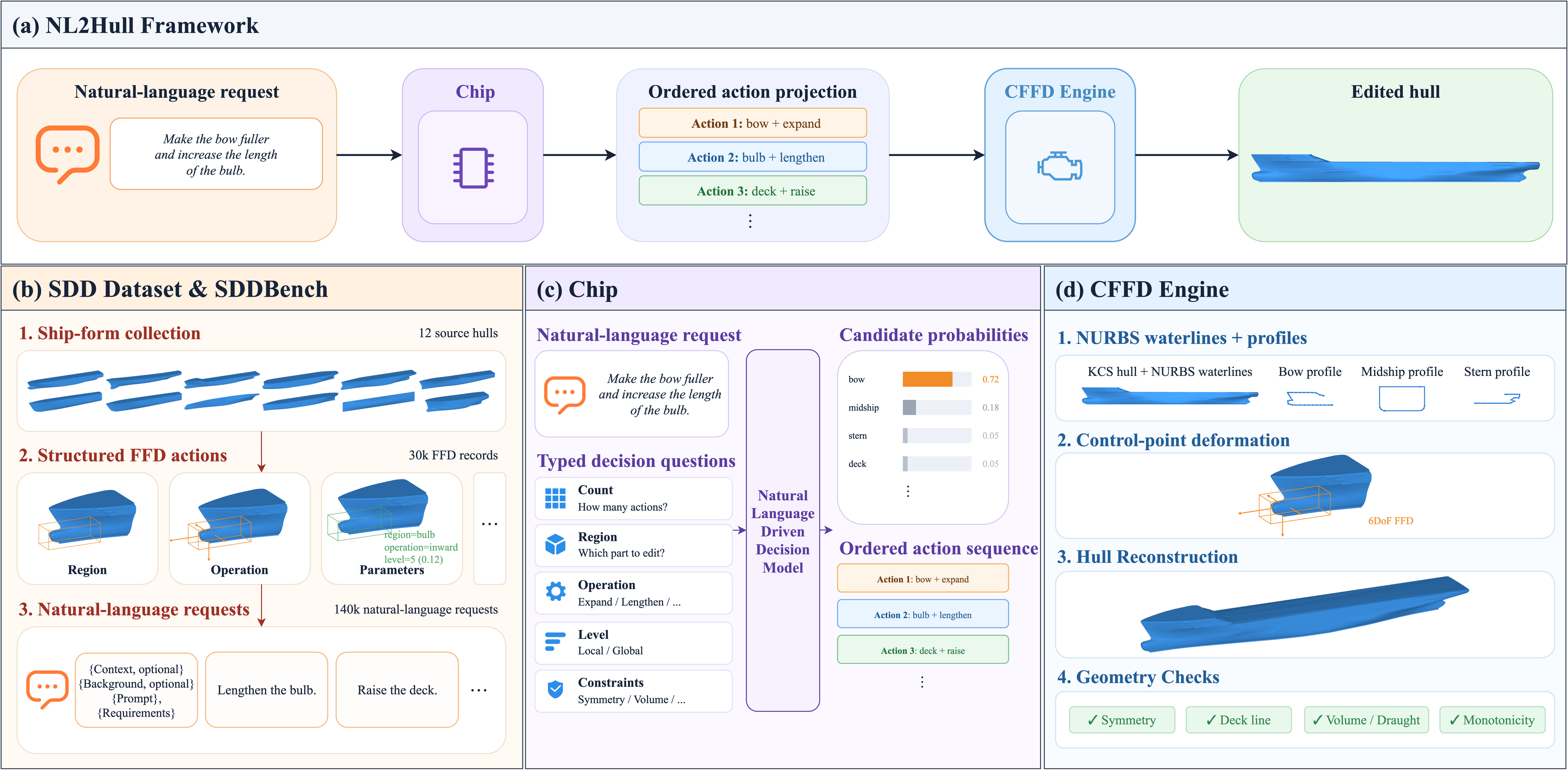}
    \caption{Overview of the NL2Hull framework and its ship-form editing workflow.
    (a) Natural-language requests are converted by Chip into ordered actions and
    applied by the CFFD Engine to a KCS benchmark hull. (b) SDD Dataset and
    SDDBench construction. (c) Typed decision questions and candidate
    probabilities in Chip. (d) NURBS representation, control-point deformation,
    hull reconstruction, and geometric checks.}
    \label{fig:nl2hull-framework}
\end{figure*}

Figure~\ref{fig:nl2hull-framework} summarizes the NL2Hull pipeline, the SDD
Dataset and SDDBench construction, Chip's typed decision interface, and the
CFFD Engine stages.

\section{Methods}
\label{sec:methods}

This section defines the CFFD Engine, the NL2Hull Framework decision interface powered by Chip,
and the SDD Dataset and SDDBench. The central design choice
is to use Chip as a typed decision layer rather than asking it to emit
unrestricted NURBS control-point coordinates.

\subsection{CFFD Engine: NURBS-Based Ship Representation and FFD}
\label{sec:nurbs-ffd}

The CFFD Engine provides a compact parameterization for localized and global
ship-form deformation. The ship form is represented by longitudinal waterline
curves and fore and aft contours in the longitudinal--vertical plane. NURBS
describe these curves through control points, positive weights, and knot vectors.
FFD then operates directly on the waterline control points, so local and global
ship-form changes can be expressed through the same parameterized representation.
Together with hull reconstruction and constraint checks, this representation and
its deformation rules constitute the CFFD Engine.

\paragraph{Coordinate system and waterline representation.}
To express different ship forms in a common coordinate system, all coordinates
are first scaled by the hull length. The longitudinal origin is at the aft
extremity, the transverse origin is on the centreline, and the vertical origin is
at the lowest point of the hull. The normalized longitudinal extent is therefore
$[0,1]$. Within this coordinate system, waterlines at prescribed elevations
describe the longitudinal distribution of half-breadth. Each half-waterline
comprises aft and fore segments joined at the midpoint of its longitudinal
extent. Each segment is represented by a cubic, clamped NURBS curve with eight
control points:
\begin{equation}
 \mathbf{c}(u)=
 \frac{\sum_{i=0}^{7} B_{i,3}(u)w_i\mathbf{p}_i}
      {\sum_{i=0}^{7} B_{i,3}(u)w_i},
 \qquad 0\leq u\leq 1,
 \label{eq:waterline-nurbs}
\end{equation}
where $\mathbf{p}_i$ is a control point, $w_i>0$ is its weight, and $B_{i,3}$ is
the cubic B-spline basis function. The knot vector is clamped at both ends and
has uniformly spaced interior knots. Both segments are parameterized from the outer hull extremity towards the
section midpoint. In this parameterization, three interior control points
coincide to describe the transition towards the central portion of the waterline.
Reflection of the half-waterline across the centreline then defines the complete
symmetric section.

\paragraph{Longitudinal profile representation.}
To describe the bow and stern contours more faithfully, the waterline curves are
complemented by fore and aft longitudinal profiles in the longitudinal--vertical
plane. These profiles cover the normalized ranges $[0.65,1]$ and $[0,0.35]$,
respectively. Each profile uses a cubic, clamped curve with 32 control points and
unit weights, following the rational curve expression in
Eq.~\ref{eq:waterline-nurbs} with the summation extended to all 32 controls. The
deck and lower hull portions within each profile share one curve representation.

To preserve local geometric detail in the fore region, its knot vector uses eight
locally concentrated interior knots around the bulb-related parameter and a knot
of multiplicity three at the upper-contour corner. The remaining interior knots
are distributed over the intervening parameter intervals, whereas the aft profile
uses uniformly spaced interior knots. At each waterline elevation, the foremost
intersection with the fore profile and the aftmost intersection with the aft
profile specify the corresponding centreline endpoints. These intersections
therefore couple the longitudinal profiles and waterlines within the ship-form
representation.

\paragraph{Control-point deformation.}
With this representation established, an editing action specifies a hull region,
an operation, a magnitude, and optional longitudinal and vertical extents. Regions
comprise the bow, stern, bulb, midbody, deck, bilge, and complete hull. The
magnitude is a fraction of the relevant hull dimension. To accommodate vague
natural-language intents, the magnitude is discretized into six discrete FFD
levels with values $0$, $0.01$, $0.02$, $0.04$, $0.08$, and $0.12$; the next
subsection describes how the decision framework selects these levels. Waterline
controls are expressed in three dimensions, with their initial vertical
coordinates set to the section elevation.

Within the selected longitudinal and vertical extents, local deformation is
weighted by each control point's position. Smooth transitions use
\begin{equation}
 S(t)=3\tau^2-2\tau^3,
 \qquad \tau=\min(1,\max(0,t)).
 \label{eq:smooth-transition}
\end{equation}
For an interval $[a,b]$ in the current normalized coordinate system, the local
weight is
\begin{equation}
 W(t;a,b)=S\!\left(\frac{t-a}{d}\right)
           S\!\left(\frac{b-t}{d}\right),
 \qquad d=0.2(b-a).
 \label{eq:local-window}
\end{equation}
Full-coordinate intervals receive unit weight, and selected terminal vertical
sections are included. The product of the longitudinal and vertical weights then
defines the local influence factor $M$.

Using this influence factor, transverse expansion and contraction multiply the
control-point half-breadth by
$1+\alpha M$ and $1-\alpha M$, respectively, where $\alpha$ is the magnitude
fraction. Fullness changes follow these transverse rules, and flare-increase
operations use the expansion rule over the selected extent. Vertical operations
translate controls by $\pm\alpha D M$, where $D$ is the current vertical extent.
Longitudinal operations translate controls by a fraction of the current hull
length, using one-sided transitions towards the bow or stern extremity and
two-sided weights in other local regions. Global length changes scale
longitudinal coordinates about the centre of the current longitudinal extent;
global breadth changes scale transverse coordinates. Actions are applied
sequentially to the updated control points.

\paragraph{Geometric constraints.}
The deformation rules are applied together with geometric constraints. Bilateral
symmetry is maintained through the half-waterline representation. Deck-line
preservation attenuates local deformation with increasing elevation and fixes the
uppermost controls. Volume-preserving edits compensate transverse control-point
coordinates using the ratio of pre-edit to post-edit enclosed volume; concurrent
deck-line preservation restricts this compensation to controls below the
uppermost level. Fixed-draught requests reject vertical operations. Finally,
longitudinal translations include an ordering correction, and updated waterlines
are checked for longitudinal monotonicity and non-negative half-breadth. An
optional shape-change threshold is imposed on changes in normalized second
differences of the longitudinal and transverse control-point coordinates.

\subsection{NL2Hull Decision Interface}
\label{sec:decision-framework}

Building on the parameterized ship form and deformation rules above, we formulate
natural-language-driven ship-form editing as a discrete decision task. A design
request is decomposed into the number of deformation actions and the region and
operation associated with each action. A natural-language decision model assigns
probabilities to predefined alternatives; the highest-probability answers then
define an ordered sequence of region--operation pairs. These decisions use the
same region and operation vocabulary as the control-point deformation rules in
Section~\ref{sec:nurbs-ffd}.

\paragraph{Decision representation.}
To make this decision process explicit, the model receives the natural-language
request together with typed questions. Each question specifies an instruction and
a set of candidate answers. The number of actions is selected from one, two, and
three. For each available action position, a region question selects among the
bow, stern, bulb, midbody, deck, bilge, and complete hull. An operation question
selects among transverse, vertical, or longitudinal movement; increased or
decreased fullness; increased flare; bulb-length modification; and increased or
decreased global length or breadth. Position-indexed questions preserve the
order of the actions in a multi-operation request.

The decision record also characterizes the requested magnitude and constraints.
A categorical question distinguishes qualitative intensity from explicit
numerical magnitude or spatial extent. A discrete FFD-level question selects
among the six levels introduced in Section~\ref{sec:nurbs-ffd}, from no change to
very large change. For requests expressed with explicit numerical values, this
question uses the no-change entry by convention. Two Boolean questions identify
requests to preserve displacement and the deck line. Together, these questions
associate the additional design requirements with their corresponding action
positions.

\paragraph{Decision model Chip.}
Chip is the ship design decision model used by NL2Hull Framework. It adapts the
Kev decision interface for natural-language ship-form editing and returns a
probability distribution over the candidate answers for each typed question
\footnote{\url{https://github.com/jaredpalmer/kev}}. The model receives a
natural-language request together with a question and its candidate answers; the
geometric state is supplied separately to the CFFD Engine during execution.

Let $t$ denote the request text, $q_j$ the $j$th typed question, and
$\mathcal{Y}_j$ its candidate set. At the interface level, Chip first maps the
request and question context to a decision representation,
\begin{equation}
 \mathbf{h}_j=f_{\theta}(t,q_j),
 \qquad
 \mathbf{s}_j=g_{\theta}(\mathbf{h}_j,\mathcal{Y}_j),
 \label{eq:chip-interface}
\end{equation}
where $f_{\theta}$ denotes language processing and $g_{\theta}$ assigns one
score to each candidate answer. The scores are converted into a probability
distribution by
\begin{equation}
 p_{\theta}(y\mid t,q_j,\mathcal{Y}_j)
 =\frac{\exp(s_{j,y})}{\sum_{y'\in\mathcal{Y}_j}\exp(s_{j,y'})},
 \qquad y\in\mathcal{Y}_j,
 \label{eq:decision-distribution}
\end{equation}
where $s_{j,y}$ is the score of candidate $y$. Categorical, discrete FFD-level,
and Boolean questions therefore share the same probability interface while using
their respective candidate sets. The selected answer is the maximum-probability
candidate,
\begin{equation}
 \hat{y}_j=\underset{y\in\mathcal{Y}_j}{\operatorname{arg\,max}}
          \ p_{\theta}(y\mid t,q_j,\mathcal{Y}_j).
 \label{eq:decision-selection}
\end{equation}
The full distributions are retained for calibration and likelihood metrics, and
the selected answers are passed to the ordered action projection. Chip is trained
with the configuration summarized in Appendix~\ref{app:training}. This interface
also supports comparison with native decision services and language models
prompted to return the same candidate probability distributions.

\paragraph{Action projection and geometric correspondence.}
After the individual decisions have been made, the selected action count
determines how many region--operation pairs are included in the decision output.
If $\hat{k}$ is the selected count, and $\hat{r}_i$ and $\hat{o}_i$ are the
selected region and operation at position $i$, the projected action sequence is
\begin{equation}
 \hat{\mathcal{A}}=
 \bigl((\hat{r}_1,\hat{o}_1),\ldots,
       (\hat{r}_{\hat{k}},\hat{o}_{\hat{k}})\bigr).
 \label{eq:action-projection}
\end{equation}
The projection checks that region and operation questions are available for every
selected action position; a count exceeding the available positions is recorded
as an invalid projection. Each projected region identifies the spatial support of
an edit, while its operation selects the corresponding control-point update rule
in Section~\ref{sec:nurbs-ffd}. For example, a bow expansion corresponds to a
transverse increase over the bow extent, whereas a global length increase
corresponds to longitudinal scaling about the hull centre. The CFFD Engine then
combines each region--operation pair with its magnitude, optional longitudinal
and vertical extents, and constraint settings before applying the complete action
sequence to the waterline control points. Thus, the projection preserves the
ordered discrete decisions made by Chip while the CFFD Engine supplies their
geometric realization.

\subsection{SDD Dataset and SDDBench}
\label{sec:dataset-benchmark}

The SDD Dataset connects ship-form representations, executable FFD actions, and
natural-language design requests. The construction process has four stages:
collecting and standardizing ship forms, generating valid structured FFD actions,
producing natural-language descriptions, and converting each description into a
typed decision record for model training and evaluation.

\begin{figure*}[t]
\centering
\includegraphics[width=0.98\textwidth]{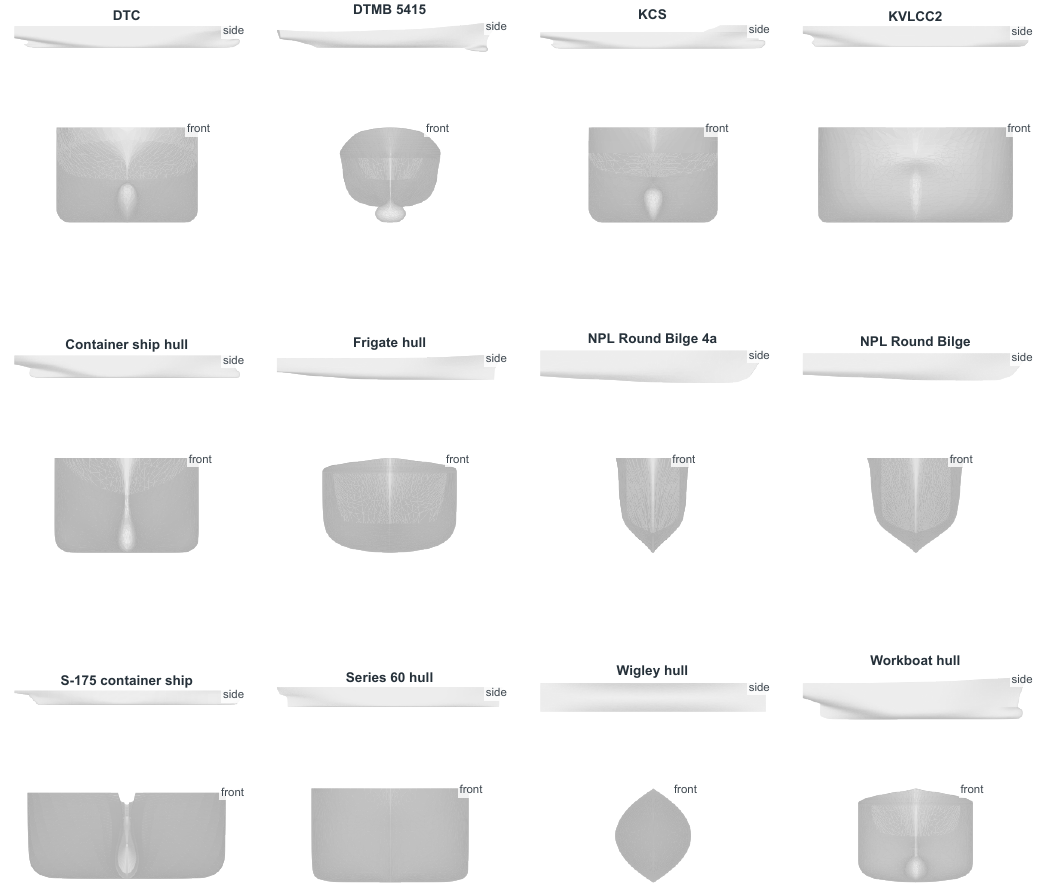}
\caption{The twelve source hull geometries used to construct the SDD Dataset.
The collection includes the Duisburg Test Case (DTC), DTMB 5415 surface-combatant
benchmark hull, KRISO Container Ship (KCS), KRISO Very Large Crude Carrier 2
(KVLCC2), a generic container-ship hull, a generic frigate hull, NPL Round Bilge
4a and NPL Round Bilge, the ITTC S-175 container ship, a Series~60 hull, the
Wigley hull, and a generic workboat hull. Side and front orthographic views use
unit-length normalization while preserving each hull's geometric aspect ratio.}
\label{fig:source-hulls}
\end{figure*}

\paragraph{Ship-form sources.}
The source collection contains twelve ship-form models. Eleven models were
collected from GrabCAD\footnote{\url{https://grabcad.com/abdelrahman.mamdouh-7}},
and the DTC model was obtained from the official OpenFOAM tutorial
resources\footnote{\url{https://github.com/OpenFOAM/OpenFOAM-7/blob/master/tutorials/resources/geometry/DTC-scaled.stl.gz}}.
The collection includes tankers, container ships, and other hull types, as shown
in Fig.~\ref{fig:source-hulls}. To use these models within a common design
framework, their coordinates were standardized with the aft extremity at
longitudinal coordinate zero, the bow at one, the baseline at vertical coordinate
zero, and the centreline at transverse coordinate zero. These standardized models
provide the geometric states for the representation and action-generation
procedures.

\paragraph{Structured FFD action dataset.}
According to the deformation rules in Section~\ref{sec:nurbs-ffd}, we randomly
sample regions and operations supported by FFD and apply each sampled action to
the waterline representation. A candidate is retained after the edited waterlines
produce a watertight solid with positive enclosed volume. Each retained record
stores the ship-form identifier, the initial state, the ordered action
specification, the interaction turns, and the geometric measurements after each
turn. The action specification includes the edited region, operation, magnitude,
spatial extents, symmetry, and optional geometric constraints.

This procedure produced 30,000 FFD operation records. These records form the
action source for the SDD Dataset and cover four design scenarios that capture
both simple edits and more involved requests. Single-edit records describe one local or
global modification and account for 9,000 records. To capture instructions that
combine several changes, a further 7,500 records specify edits to two or three
regions in a single request.

Iterative design also requires a modification to be refined in a later step.
Accordingly, 7,500 records describe two successive edits in the same region,
with the second edit continuing the previous change or, for directional
operations, acting in the opposite direction. Finally, 6,000 records express
the desired change through an explicit numerical magnitude and longitudinal and
vertical ranges. Together, these scenarios provide examples of isolated edits,
combined edits, sequential refinement, and numerically specified requests.

\paragraph{Natural-language description generation.}
Each FFD operation record provides the basis for generating Chinese
natural-language design requests. Its structured action specification is given
through the Pi Agent interface\footnote{\url{https://pi.dev}} to one of seven
language models: Claude Opus 5~\citep{anthropic2026claude}, Kimi K3~\citep{kimi2026k3},
GLM-5.3~\citep{glm5team2026glm5vibecodingagentic}, Grok-4.7,
DeepSeek-V4.1-Flash~\citep{deepseek2026v41flash}, GPT-5.6 Sol, and
Xiaomi MiMo-V2.6-Pro~\citep{mimo2026v26pro}. Four alternative
descriptions are requested for each editing step, including both steps in a
two-step record. The generation prompt is provided in Appendix~\ref{app:description-prompt}.
The generation instruction requires these descriptions to
preserve the edited region, operation, relative magnitude, spatial range, and
geometric constraints. For numerically specified requests, the numerical values
must be reproduced exactly.

To increase data diversity, we vary both the source of the descriptions and the
way the design intent is expressed. Distributing generation across seven models
reduces dependence on a single model's phrasing. Within each editing step,
multiple descriptions vary sentence structure, verbs, and intensity expressions
while referring to the same target action. Qualitative requests use expressions
such as slight or substantial change, whereas precise requests retain the
specified numerical magnitude and ranges. In addition, a subset of the
requests includes a short ship-design or engineering background. This introduces
contextual variation around the same editing goal, alongside requests that state
the modification directly.

Before constructing the typed decision records, the generated descriptions are
checked for plain-language format, required numerical values, and disclosure of
internal magnitude labels. The original natural-language collection contains
149,998 descriptions. We exclude 8,076 rollback-oriented augmentations with
invalid labels, then remove 7,364 exact text duplicates using split priority. Identical texts with conflicting labels are
rejected during validation. Thus, the resulting SDD Dataset contains 134,558
records. The training, validation, and test partitions contain 88,604, 22,758,
and 23,196 records, respectively.

Each natural-language record is expanded into typed questions with labelled
candidate answers. The first question selects one, two, or three actions. Each
action position then receives questions for its region and operation, followed by
questions for qualitative or explicit magnitude mode, the discrete FFD level,
displacement preservation, and deck-line preservation. Categorical questions use
named alternatives, discrete FFD-level questions use six levels from no change to
very large change, and Boolean questions use true or false alternatives. The
original action order is retained in the question identifiers and labels.

\begin{figure*}[t]
    \centering
    \begin{subfigure}[t]{0.63\textwidth}
        \centering
        \includegraphics[width=\linewidth]{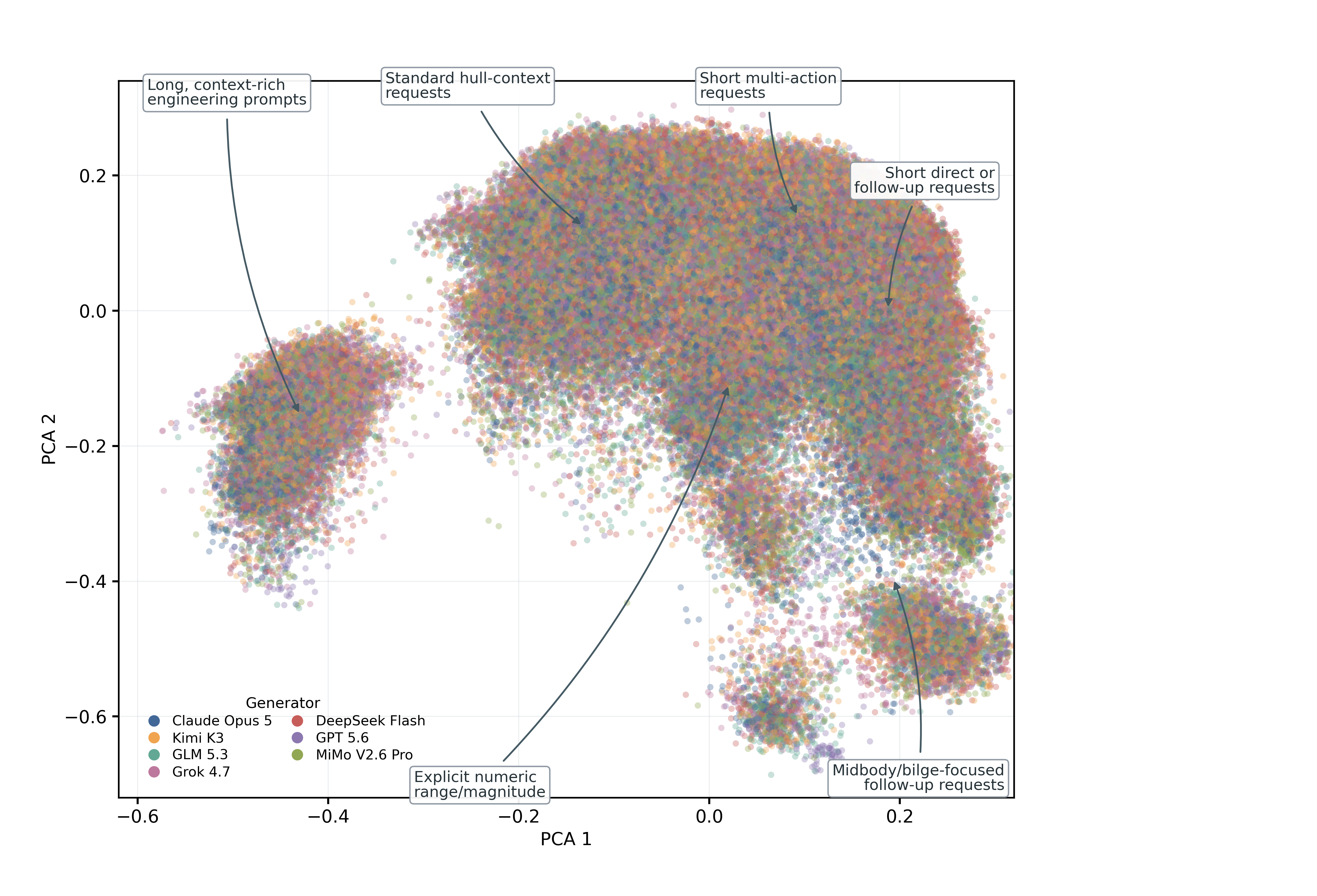}
        \caption{Qwen3-Embedding~\citep{zhang2025qwen3embedding} PCA projection.}
    \end{subfigure}\hfill
    \begin{subfigure}[t]{0.34\textwidth}
        \centering
        \includegraphics[width=\linewidth]{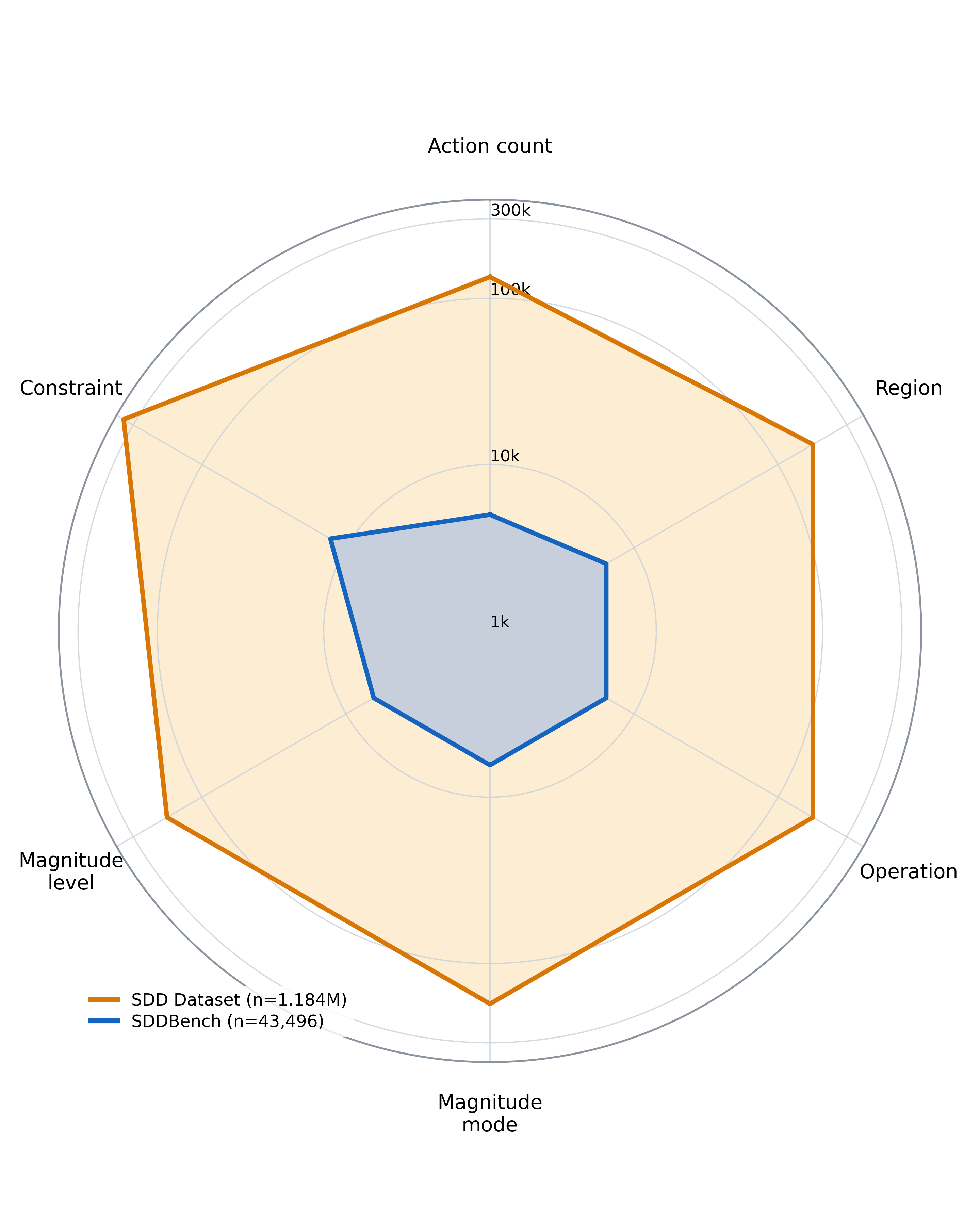}
        \caption{Typed-question counts.}
    \end{subfigure}
    \caption{Dataset structure and typed-question composition. (a) The distribution
    of the 134,558 language descriptions after embedding projection and dimensionality
    reduction. The six annotations identify dominant linguistic or structural patterns
    obtained by joining the projection to the structured records;
    (b) Absolute counts of the six typed-question families in the SDD Dataset
    and SDDBench on a common radial scale. The outer polygon is the full SDD
    Dataset (1,184,336 questions), and the inner polygon is SDDBench (43,496
    questions).}
    \label{fig:dataset-distribution}
\end{figure*}

The test partition contains 23,196 records and 201,774 expanded questions. We
select 5,000 records by hull, description source, action count, and the presence
of an explicit magnitude to form SDDBench for rapid model evaluation. The
selected subset contains 43,496 expanded questions, and every compared system
receives the same selected requests. Figure~\ref{fig:dataset-distribution}
visualizes the language-space structure and the typed-question composition of the
two collections. The near-overlap of the two radar profiles shows that SDDBench
preserves the question-family composition of the SDD Dataset while reducing its
scale for rapid evaluation.

\paragraph{Evaluation protocol and metrics.}
All compared systems receive the same SDDBench records, typed questions,
candidate sets, output format, probability validation, and metric aggregation procedure.
For each question, the model returns a probability distribution over all listed
answers. Question-level performance is measured by clean accuracy, negative
log-likelihood (NLL), expected calibration error (ECE), and Brier score. These
probability metrics follow established calibration practice~\citep{guo2017calibration,kull2019dirichlet}.
They use the expanded question set and retain the full probability distributions.

The maximum-probability answers are also projected into an ordered sequence of
region--operation pairs. Free-form deformation action accuracy is computed from
three components: the selected number of actions, the region at each action
position, and the operation at each action position. A record is counted as an
exact action match when the number, order, regions, and operations all agree with
the labelled sequence. The turn-level denominator contains all 5,000 requested
records. Model-call failures, malformed responses, invalid probabilities, missing
answers, and incompatible predicted action counts are counted as failed records
in this denominator. This protocol reports probability-based decision quality and
action-level FFD quality as separate measurements.

\section{Results}
\label{sec:results}

\subsection{NURBS Representation and FFD Accuracy}
\label{sec:nurbs-results}

This section evaluates the ship-form representation and design workflow. First,
the NURBS representation is assessed by reconstructing ship hulls from their
extracted waterlines and longitudinal profiles. Second, the FFD operator is
assessed through representative local, global, and shape-oriented edits applied
to a fitted hull. The first experiment quantifies reconstruction error across a
diverse hull set, while the second combines visual inspection with geometric
validity and volume measurements.

\subsubsection{NURBS-based hull reconstruction}
\label{sec:nurbs-reconstruction-results}

The reconstruction experiment used twelve triangulated ship hulls covering
container ships, tankers, naval hulls, benchmark hulls, and other geometries. For
each input, the procedure extracted 33 horizontal sections, fitted the aft and fore
segments of each section with cubic clamped NURBS curves, fitted the aft and fore
longitudinal profiles, and reconstructed a closed hull mesh by connecting the
sampled section rings. The reconstructed mesh was compared with the input surface
using nearest-neighbour distances. Additional errors were calculated for waterline
half-breadth and for the two longitudinal profile fits. All coordinates were
normalized by the input hull length, so the reported errors are dimensionless.

\begin{figure*}[t]
    \centering
    \begin{subfigure}[t]{0.98\textwidth}
        \centering
        \includegraphics[width=\linewidth]{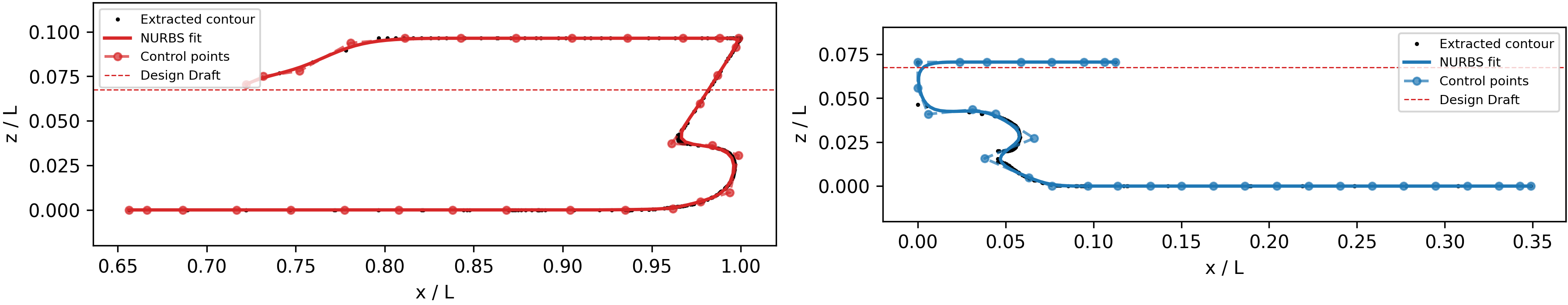}
        \caption{Longitudinal profile fitting.}
    \end{subfigure}
    \par\medskip
    \begin{subfigure}[t]{0.48\textwidth}
        \centering
        \includegraphics[width=\linewidth]{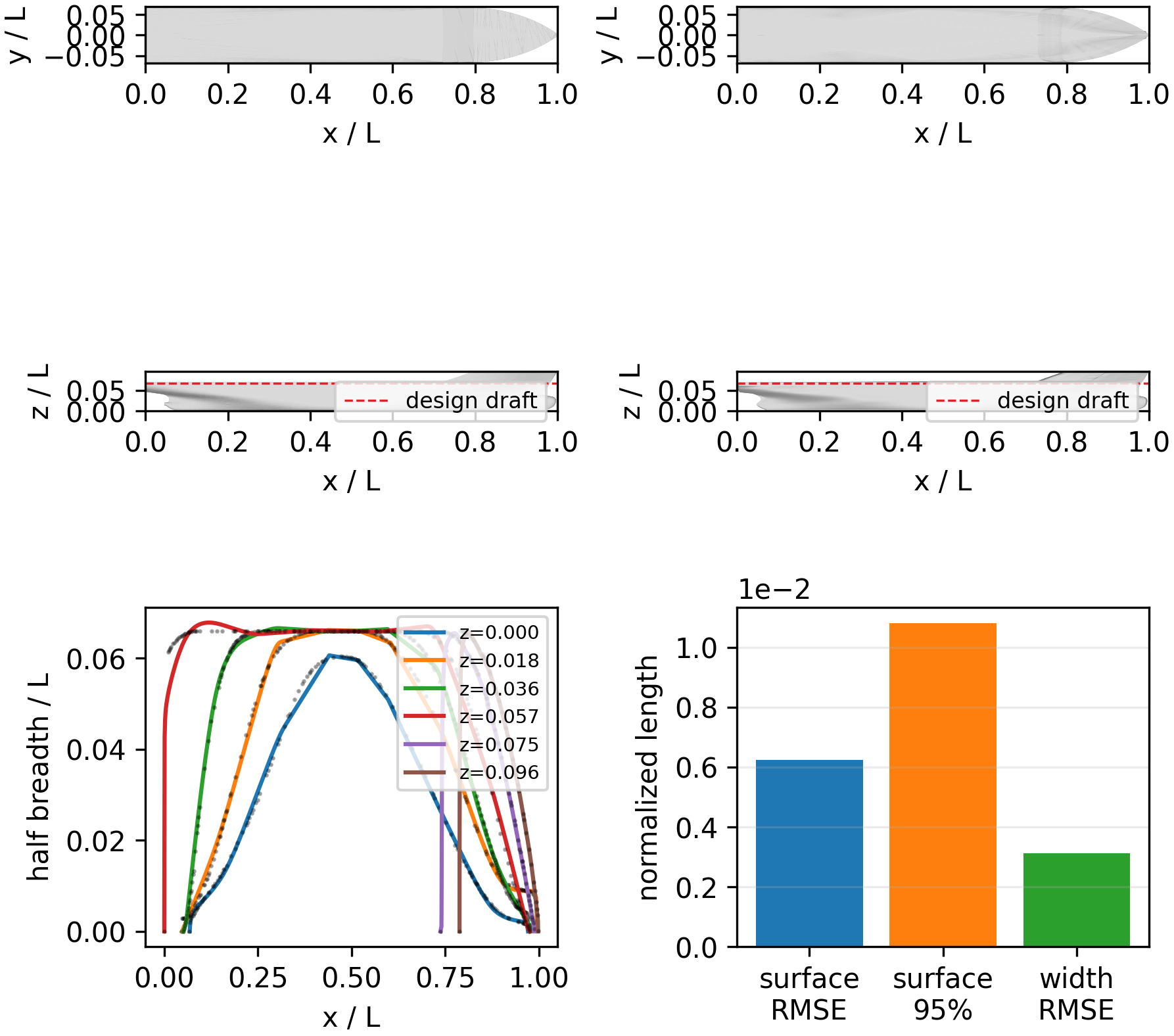}
        \caption{Hull reconstruction comparison.}
    \end{subfigure}\hfill
    \begin{subfigure}[t]{0.48\textwidth}
        \centering
        \includegraphics[width=\linewidth]{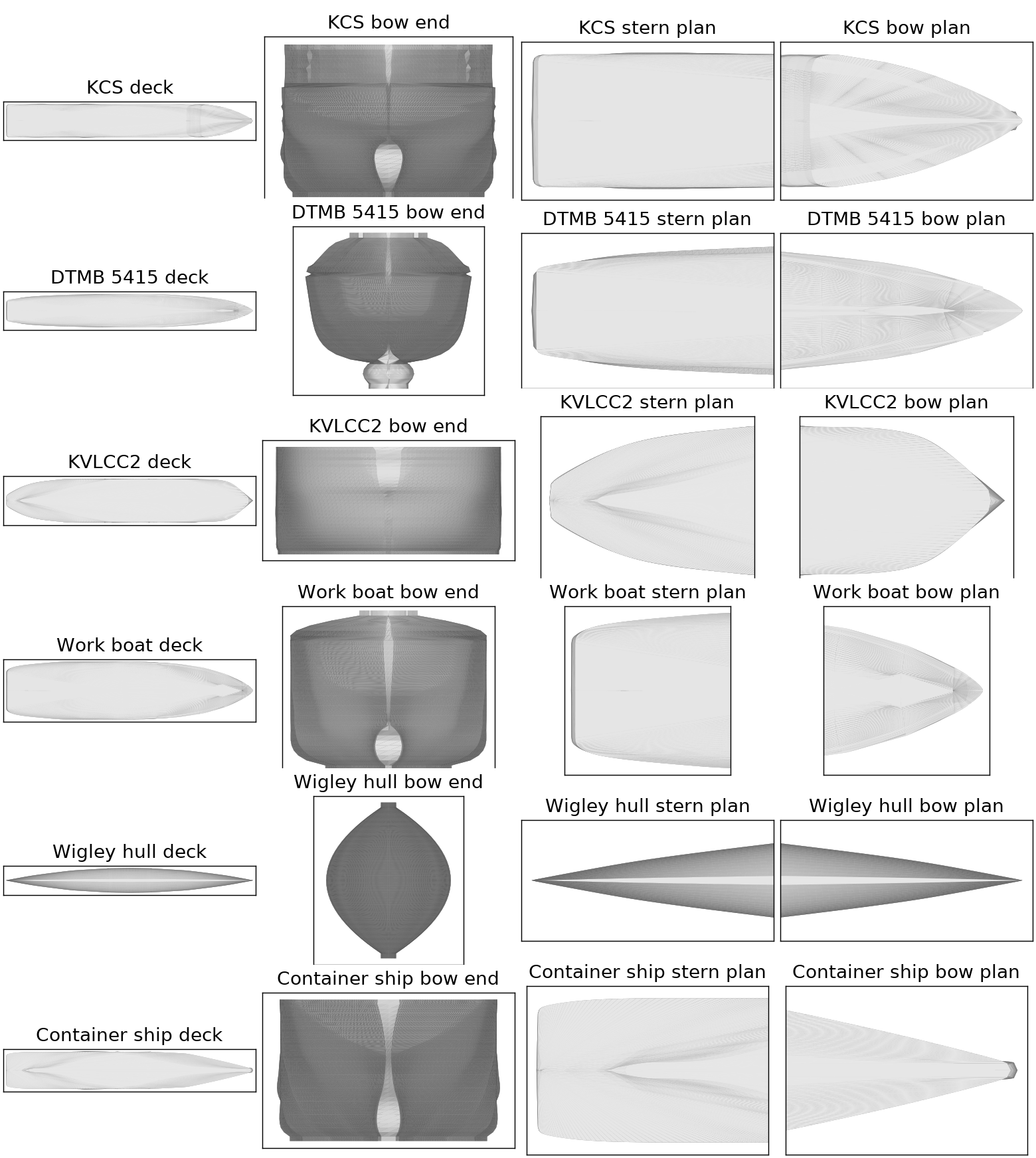}
        \caption{Closure diagnostic.}
    \end{subfigure}
    \caption{NURBS-based reconstruction of the KCS hull. The input surface is
    represented by fitted longitudinal profiles and waterline curves, then
    reconstructed as a closed section-based hull mesh. The wide profile panel is
    placed above the two compact diagnostics to preserve the aspect ratio of each
    source plot.}
    \label{fig:kcs-reconstruction}
\end{figure*}

Figure~\ref{fig:kcs-reconstruction} illustrates the reconstruction for the KCS
hull. The longitudinal profile plot shows the fitted fore and aft curves against
the extracted centre-plane contours. The three-dimensional comparison shows the
input and reconstructed hull surfaces, while the closure diagnostic records the
geometric closure of the reconstructed section-based surface.

Table~\ref{tab:reconstruction-error} summarizes the reconstruction errors. The
surface nearest-neighbour root-mean-square error (RMSE) ranged from 0.0024 for
Wigley to 0.0113 for DTC. KCS achieved a surface nearest-neighbour RMSE of
0.0062, a waterline half-breadth RMSE of 0.0031, a fore-profile RMSE of 0.0003,
and an aft-profile RMSE of 0.0008. Across the twelve hulls, fore and aft
profile-fitting errors remained below 0.0013 when measured by RMSE. These measurements show that the parameterization preserves
the extracted longitudinal and transverse geometry across multiple hull forms.

\begin{table*}[t]
\caption{Reconstruction errors for the twelve-hull NURBS representation
experiment. All quantities are normalized by the input hull length. Surface RMSE
is computed from bidirectional nearest-neighbour distances between input and
reconstructed surface vertices.}
\label{tab:reconstruction-error}
\centering
\begin{tabular*}{\tblwidth}{@{\extracolsep{\fill}}lrrrr@{}}
\toprule
Hull & Surface RMSE & Waterline width RMSE & Fore-profile RMSE & Aft-profile RMSE \\
\midrule
DTC                         & 0.01130 & 0.00302 & 0.00038 & 0.00121 \\
DTMB5415                    & 0.00524 & 0.00227 & 0.00070 & 0.00072 \\
KCS                         & 0.00623 & 0.00312 & 0.00034 & 0.00082 \\
KVLCC2                      & 0.00975 & 0.00140 & 0.00012 & 0.00092 \\
Containership               & 0.00690 & 0.00202 & 0.00003 & 0.00038 \\
Firgate                     & 0.00658 & 0.00284 & 0.00023 & 0.00084 \\
NPL round bilge (model)     & 0.00849 & 0.00338 & 0.00004 & 0.00052 \\
NPL round bilge (full scale)& 0.00546 & 0.00208 & 0.00008 & 0.00097 \\
S-175 U-water               & 0.01006 & 0.00313 & 0.00012 & 0.00079 \\
Series 60                   & 0.00893 & 0.00094 & 0.00021 & 0.00055 \\
Wigley hull                 & 0.00241 & 0.00018 & 0.00011 & 0.00079 \\
Work boat                   & 0.00761 & 0.00504 & 0.00030 & 0.00090 \\
\bottomrule
\end{tabular*}
\end{table*}

\subsubsection{FFD operation visualization and geometric validity}
\label{sec:ffd-results}

The FFD experiment applied 30 predefined operations to a fitted KVLCC2 hull. The
cases covered bow, stern, and bulb edits; midbody, deck, and bilge edits; global
length and breadth changes; and shape-oriented operations such as fullness, flare,
and bulb-length changes. Directional operations used the plan view for transverse
changes and the side view for vertical and longitudinal changes. Each deformed hull
was reconstructed from the updated NURBS waterlines and evaluated for watertightness
and valid enclosed volume.

\begin{figure*}[t]
    \centering
    \begin{subfigure}[t]{0.98\textwidth}
        \centering
        \includegraphics[width=\linewidth]{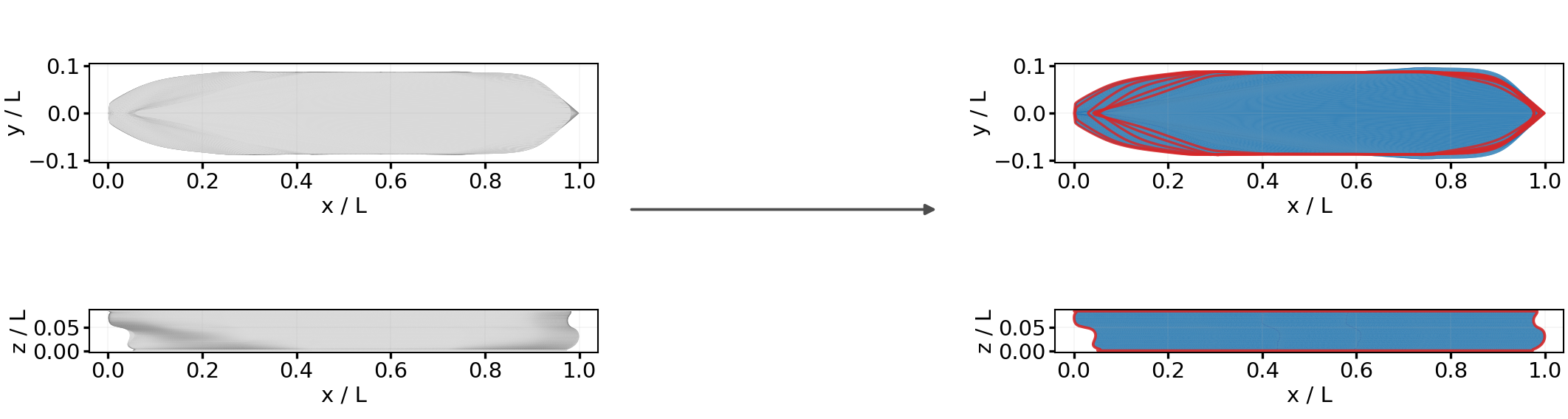}
        \caption{Bow outward.}
    \end{subfigure}
    \par\smallskip
    \begin{subfigure}[t]{0.98\textwidth}
        \centering
        \includegraphics[width=\linewidth]{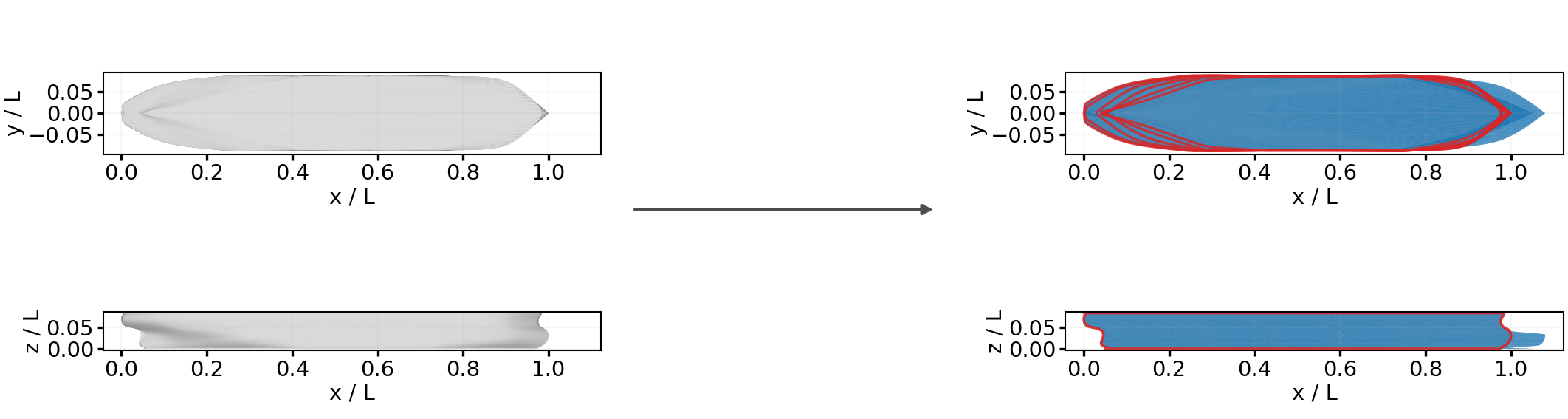}
        \caption{Bulb forward.}
    \end{subfigure}
    \par\smallskip
    \begin{subfigure}[t]{0.98\textwidth}
        \centering
        \includegraphics[width=\linewidth]{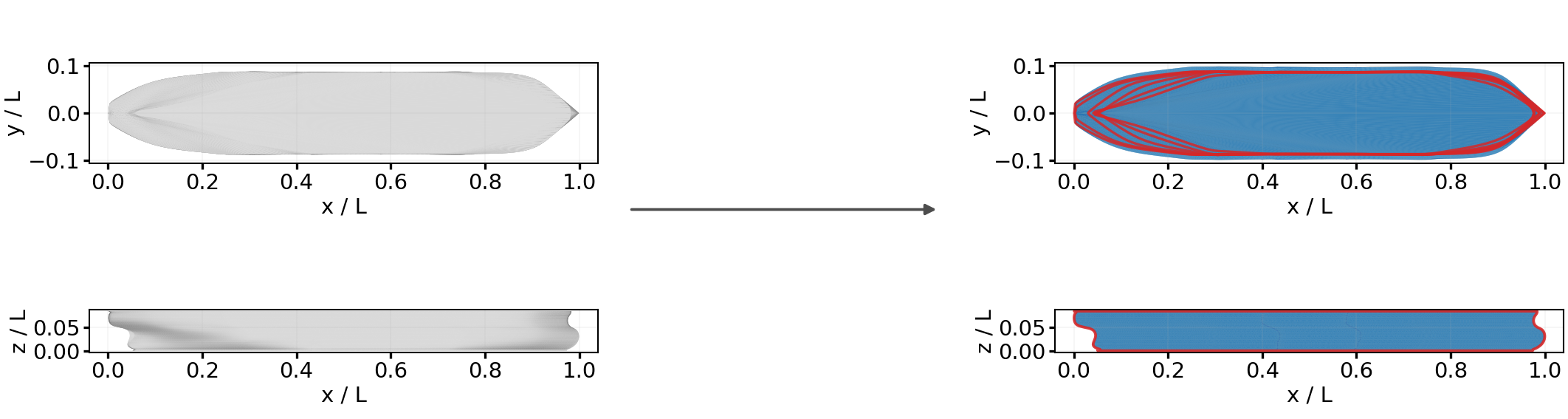}
        \caption{Global breadth increase.}
    \end{subfigure}
    \caption{Representative FFD operations applied to the fitted KVLCC2 hull.
    Each panel compares the original fitted hull with the hull obtained after the
    indicated control-point deformation. In each row, the centered arrow points
    from the original hull to the modified hull. The right column overlays the
    deformed hull (blue) with red section lines from the original hull (gray);
    plan views are above side views.}
    \label{fig:ffd-representative}
\end{figure*}

Figure~\ref{fig:ffd-representative} presents three representative operations. The
bow-outward case expands the forward transverse form, the bulb-forward case moves
the bulb region in the longitudinal direction, and the global-breadth case scales
the transverse dimension across the hull. The displayed changes remain localized
for the first two operations and span the complete hull for the third operation.

Table~\ref{tab:ffd-validity} reports the geometric measurements for the same
representative cases. The bow-outward, bulb-forward, and global-breadth operations
produced volume ratios of 1.037, 1.033, and 1.120, respectively. All 30 tested
cases passed the watertightness and enclosed-volume checks. The volume changes
follow the selected operation: local transverse and longitudinal edits modify the
enclosed volume in the affected region, whereas global breadth scaling produces a
larger volume change across the entire hull.

\begin{table}[t]
\caption{Geometric measurements for representative FFD operations on KVLCC2.}
\label{tab:ffd-validity}
\centering
\begin{tabular*}{\tblwidth}{@{\extracolsep{\fill}}lrrcc@{}}
\toprule
Operation & Level & \shortstack{Volume\\ratio} & Watertight & \shortstack{Valid\\volume} \\
\midrule
Bow outward             & 5 & 1.037 & Yes & Yes \\
Bulb forward            & 4 & 1.033 & Yes & Yes \\
\shortstack[l]{Global breadth\\increase} & 5 & 1.120 & Yes & Yes \\
\bottomrule
\end{tabular*}
\end{table}

Together, the reconstruction and deformation experiments establish a continuous
geometric workflow from stereolithography (STL) input to NURBS parameters, from control-point updates
to reconstructed hull surfaces, and from local or global FFD operations to
geometric validity checks. The reconstruction errors quantify representation
fidelity, while the FFD cases demonstrate the spatial selectivity and global
scaling behaviour of the deformation operators. These experiments demonstrate
the feasibility of the ship-form representation and reconstruction pipeline,
providing the CFFD Engine for subsequent natural-language-driven ship
design.

\subsection{SDDBench Evaluation}
\label{sec:benchmark}

\subsubsection{Performance on SDDBench}
\label{sec:ship-decision-results}

To evaluate model performance in ship design decision making, all models were
tested on SDDBench. The comparison includes Chip, the native JEV decision model,
three flagship language models, and three locally served small language models:
Qwen3~\citep{yang2025qwen3}, Gemma 3~\citep{gemmateam2025gemma3}, and Llama 3.2~\citep{grattafiori2024llama3}.
Chip and JEV expose typed decision interfaces directly. For the general language
models, we used prompt-constrained adaptation: each model received the record
state, typed questions, candidate keys, and a strict JavaScript Object Notation
(JSON) output schema, and was
instructed to return a probability distribution for every question. The resulting
responses were parsed and validated before the common action projection and
metric aggregation. The general language models were invoked through Pi
Agent\footnote{\url{https://pi.dev}} in non-interactive mode, with agent tools
and persistent sessions disabled. The prompt is given in
Appendix~\ref{app:benchmark-prompt}.

Question-level metrics are computed over successfully evaluated questions. For
each question $i$, the model produces a probability distribution $p_i(c)$ over
the labelled candidate answers $c\in\mathcal{Y}_i$, with true answer $y_i$.
The question accuracy measures the fraction of questions whose most probable
answer is correct. NLL is
\begin{equation}
\mathrm{NLL}=-\frac{1}{N}\sum_{i=1}^{N}\log p_i(y_i),
\label{eq:nll}
\end{equation}
where $N$ is the number of successfully evaluated questions. For multiclass
candidate distributions, the Brier score is
\begin{equation}
\mathrm{Brier}=\frac{1}{N}\sum_{i=1}^{N}\sum_{c\in\mathcal{Y}_i}
\left[p_i(c)-\mathbb{1}(c=y_i)\right]^2.
\label{eq:brier}
\end{equation}
To calculate ECE, the questions are grouped into
confidence bins $B_m$ using $\hat{p}_i=\max_c p_i(c)$. With
$\mathrm{acc}(B_m)$ denoting the fraction of correct argmax predictions and
$\mathrm{conf}(B_m)$ the mean confidence in bin $m$,
\begin{equation}
\mathrm{ECE}=\sum_{m=1}^{M}\frac{|B_m|}{N}
\left|\mathrm{acc}(B_m)-\mathrm{conf}(B_m)\right|.
\label{eq:ece}
\end{equation}
Here $\mathbb{1}$ denotes the indicator function, and $M=10$ equal-width
confidence bins are used; empty bins contribute zero to ECE. The Brier score
uses one-hot targets for every question type, including discrete magnitude
levels. FFD exact match is computed after projecting the most probable typed
answers into an
ordered sequence of region--operation pairs. Its denominator contains all 5,000
requested records, including failed calls, malformed outputs, invalid probability
structures, and incompatible predicted action counts. Table~\ref{tab:benchmark-outline}
summarizes the comparison.

\begin{table*}[t]
\caption{Performance on SDDBench. Question accuracy is computed over successfully
evaluated questions, and FFD exact match uses all 5,000 requested records. Accuracy values are percentages; lower values
are preferred for NLL, ECE, and Brier score. Best values are shown in bold.}
\label{tab:benchmark-outline}
\centering
\begin{tabular*}{\tblwidth}{@{\extracolsep{\fill}}lrrrrr@{}}
\toprule
Model & Question accuracy $\uparrow$ & FFD exact match $\uparrow$ & NLL $\downarrow$ & ECE $\downarrow$ & Brier $\downarrow$ \\
\midrule
\multicolumn{6}{l}{\textit{Small language models}} \\
Qwen3                 & 75.61 & 2.84  & 1.2088 & 0.1148 & 0.3803 \\
Gemma3                & 66.97 & 15.98 & 1.4328 & 0.0225 & 0.4471 \\
Llama3.2              & 63.94 & 2.16  & 1.4999 & 0.0723 & 0.5063 \\
\hline
\multicolumn{6}{l}{\textit{Flagship language models}} \\
Claude Opus 5         & 92.87 & 99.12 & 0.2385 & 0.0607 & 0.1066 \\
GPT-5.6 Sol           & 92.44 & 98.32 & 0.5779 & 0.0571 & 0.1340 \\
DeepSeek V4.1 Flash   & 92.84 & 98.40 & 0.6808 & 0.0307 & 0.1158 \\
\hline
\multicolumn{6}{l}{\textit{Language decision models}} \\
JEV                   & 90.96 & 84.52 & 0.3092 & 0.0271 & 0.1266 \\
Chip (Ours)       & \textbf{95.90} & \textbf{99.32} & \textbf{0.0951} & \textbf{0.0032} & \textbf{0.0551} \\
\bottomrule
\end{tabular*}
\end{table*}

Chip achieved the highest question accuracy among the compared systems,
reaching 95.90\%, and also achieved the highest FFD exact match at
99.32\%. Its NLL, ECE, and Brier score were 0.0951, 0.0032, and 0.0551, respectively. These values were lower than the
corresponding values of JEV, which reached 90.96\% question accuracy and 84.52\%
FFD exact match. Chip therefore combined accurate typed answers
with well-calibrated probability outputs on SDDBench.

The flagship language models produced competitive action-level results. Claude
Opus 5 reached 99.12\% FFD exact match and GPT-5.6 Sol reached 98.32\%, while
their question accuracies were 92.87\% and 92.44\%. DeepSeek V4.1 Flash
reached 92.84\% question accuracy and 98.40\% FFD exact match. The action-level
scores of these models were close to Chip, while their
question-level accuracies were lower and their probability error measures were
higher than those of Chip.

The locally served models showed a different pattern under the same output
protocol. Qwen3, Gemma3, and Llama3.2 achieved question accuracies of 75.61\%,
66.97\%, and 63.94\% on their successfully evaluated questions. Their FFD
exact-match scores were 2.84\%, 15.98\%, and 2.16\%, respectively. Invalid or
failed probability responses occurred in 4,225, 2,359, and 4,443 records for these
three models, respectively, compared with 678 for JEV and 31
for DeepSeek. Chip, Claude Opus 5, and GPT-5.6 Sol returned valid probability
responses for all 5,000 records. Incompatible predicted action counts were also
counted as FFD projection failures. Because FFD exact match retains every
requested record in its denominator, these results jointly
measure decision quality and the reliability of producing a valid typed response
under the evaluation interface.

\subsubsection{General decision capability on JevBench}
\label{sec:jevbench-results}

To assess general decision performance after training on the ship-design
SDD Dataset, Chip was tested on the public JevBench
benchmark\footnote{\url{https://github.com/fstandhartinger/jevbench}}.
This evaluation applies the same typed-decision interface to tasks outside ship
design. The Easy, Original, and Hard tiers contained 48, 72, and 111 planned tasks,
respectively. All planned tasks produced valid probability distributions and were
scorable under the public-only protocol.

Chip solved all 48 Easy tasks, solved 59 of 72 Original tasks, and solved 41
of 111 Hard tasks. The corresponding accuracies were 100.0000\%, 81.9444\%, and
36.9369\%. The Original tier produced a Brier score of 0.3489 and an expected
calibration error of 0.1633. The Easy and Hard tiers produced Brier scores of
0.0026 and 0.8782, with ECE values of 0.0075 and 0.3577,
respectively. These results show that Chip remains operational on external typed-decision tasks, while task difficulty produces a
clear performance gradient.

Table~\ref{tab:jevbench-results} presents the three tier accuracies alongside
public reference results for other typed-decision models.

\begin{table*}[t]
\caption{General typed-decision evaluation on JevBench. Accuracy values are
percentages. Brier score and ECE are computed by the
corresponding benchmark scorers; lower values are preferred. Chip calibration
metrics are from the Original tier. Public reference calibration values follow
the reporting scope of the supplied JevBench comparison table.}
\label{tab:jevbench-results}
\centering
\begin{tabular*}{\tblwidth}{@{\extracolsep{\fill}}lrrrrr@{}}
\toprule
Model & Easy & Original & Hard & Brier $\downarrow$ & ECE $\downarrow$ \\
\midrule
JEV                  & \textbf{100.0000} & \textbf{98.6100} & 72.0700 & 0.3584 & 0.0947 \\
Laya                 & 95.8300  & 72.2200 & 28.8300 & 0.8042 & 0.2465 \\
SemIf                & \textbf{100.0000} & \textbf{98.6100} & 61.2600 & 0.4980 & 0.1122 \\
JevK5                & \textbf{100.0000} & 97.2200 & \textbf{73.8700} & 0.3662 & \textbf{0.0467} \\
Intern-Decision-0.8B & 97.9200  & 80.5600 & 52.2500 & 0.5295 & 0.0657 \\
Chip (Ours)      & \textbf{100.0000} & 81.9444 & 36.9369 & \textbf{0.3489} & 0.1633 \\
\bottomrule
\end{tabular*}
\end{table*}

On the Easy tier, Chip matched the best accuracy in the supplied comparison.
Its Original-tier accuracy exceeded the reported values for Laya and
Intern-Decision-0.8B, while the Hard-tier accuracy remained below the strongest
public reference rows. The increasing probability errors from the Easy tier to
the Hard tier accompany this decline in accuracy. Taken together, the JevBench
evaluation confirms
that ship-design specialization preserves a functioning typed-decision capability
on an external benchmark, with performance determined by the difficulty of the
underlying decision tasks.

\subsection{End-to-End Ship Form Editing Evaluation}
\label{sec:end-to-end}

This section evaluates the end-to-end editing stage of NL2Hull Framework and its
computational time. The workflow receives a natural-language instruction, predicts
typed deformation decisions, projects them into executable FFD actions, updates
the NURBS waterline parameters through the CFFD Engine, reconstructs the edited
hull, and checks geometric validity and design constraints. The evaluation therefore
connects decision accuracy with the resulting ship-form edit.

\subsubsection{End-to-end ship-form editing}
\label{sec:end-to-end-editing}

The end-to-end evaluation used the KVLCC2 hull and 300 sampled experiments from
the SDD Dataset. The experiments were evenly divided among single edits,
multi-region edits within one turn, and two consecutive edits applied to the same
region. The 300 experiments contained 400 natural-language requests because each
continuous two-turn experiment included two requests. The evaluation used
qualitative magnitude requests, while requests requiring direct continuous
magnitude or spatial-extent prediction were excluded from this interface.

For each request, the model output was converted into an executable deformation
action. The actions were applied in sequence to the current NURBS waterline state.
The resulting waterlines were lofted into a hull mesh, after which the evaluator
checked mesh validity, constraint satisfaction, action order, and qualitative
magnitude levels. An experiment was counted as successful when the predicted
trajectory matched the labelled action sequence and the reconstructed hull passed
the geometric and constraint checks.

\begin{figure*}[t]
    \centering
    \begin{subfigure}[t]{0.98\textwidth}
        \centering
        \includegraphics[width=\linewidth]{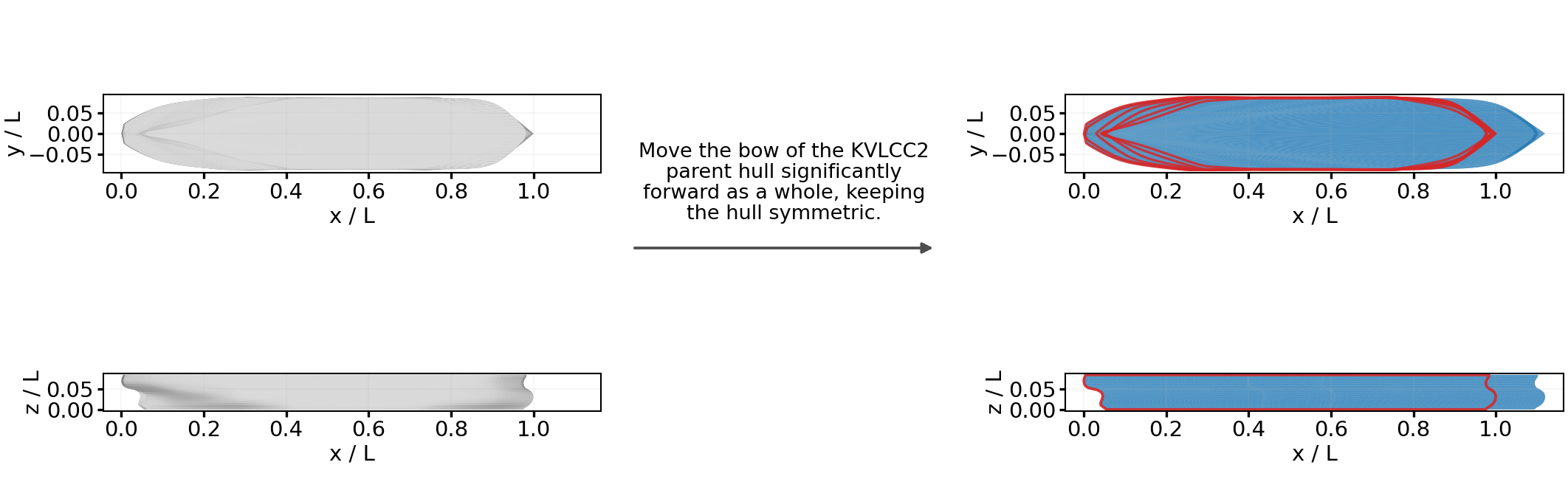}
        \caption{Single edit.}
    \end{subfigure}
    \par\smallskip
    \begin{subfigure}[t]{0.98\textwidth}
        \centering
        \includegraphics[width=\linewidth]{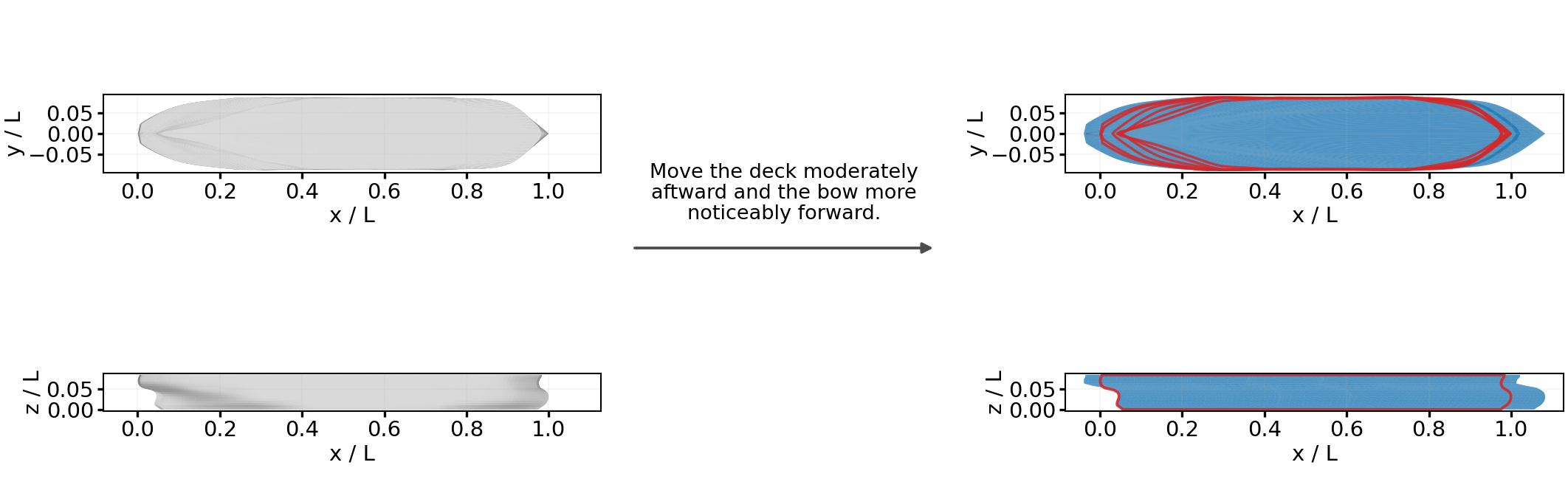}
        \caption{Multi-action edit.}
    \end{subfigure}
    \par\smallskip
    \begin{subfigure}[t]{0.98\textwidth}
        \centering
        \includegraphics[width=\linewidth]{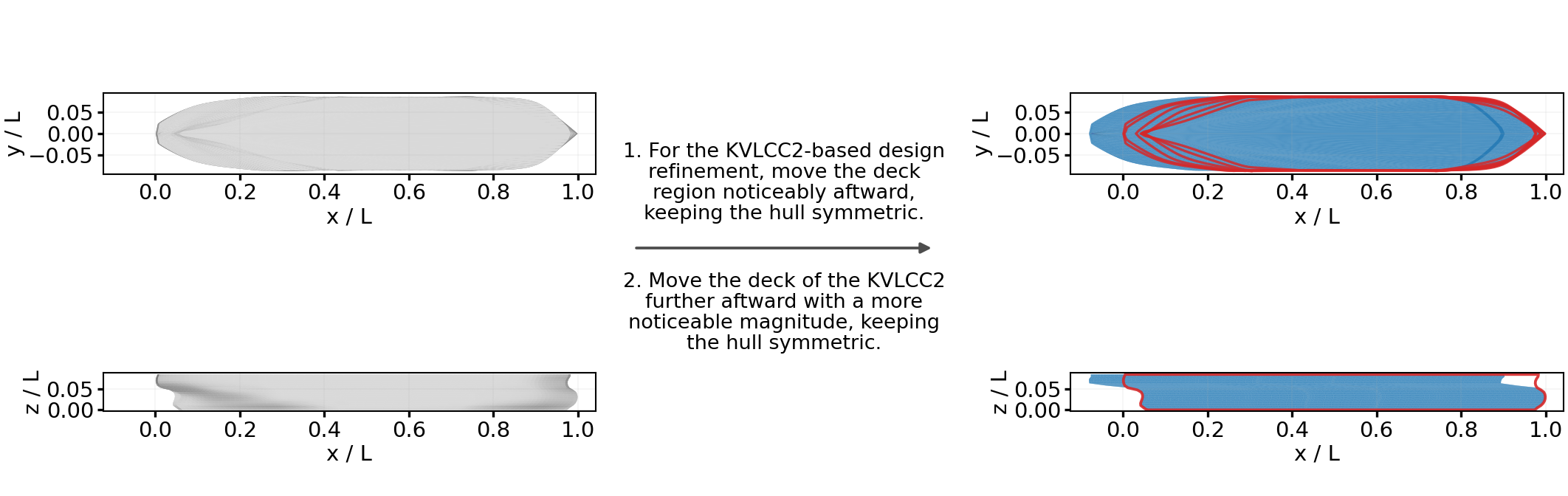}
        \caption{Continuous two-turn edit.}
    \end{subfigure}
    \caption{Representative end-to-end ship-form editing cases. Each panel
    connects a natural-language instruction to the original and modified KVLCC2
    hulls in plan and side views. Each interaction category occupies one row,
    with the original hull on the left and the modified hull (blue) on the right;
    the centered arrow indicates the editing direction, and red curves trace
    sections of the original hull. The displayed instructions
    are English translations of the evaluated Chinese requests.}
    \label{fig:end2end-primary}
\end{figure*}

\begin{figure*}[t]
    \centering
    \begin{subfigure}[t]{0.98\textwidth}
        \centering
        \includegraphics[width=\linewidth]{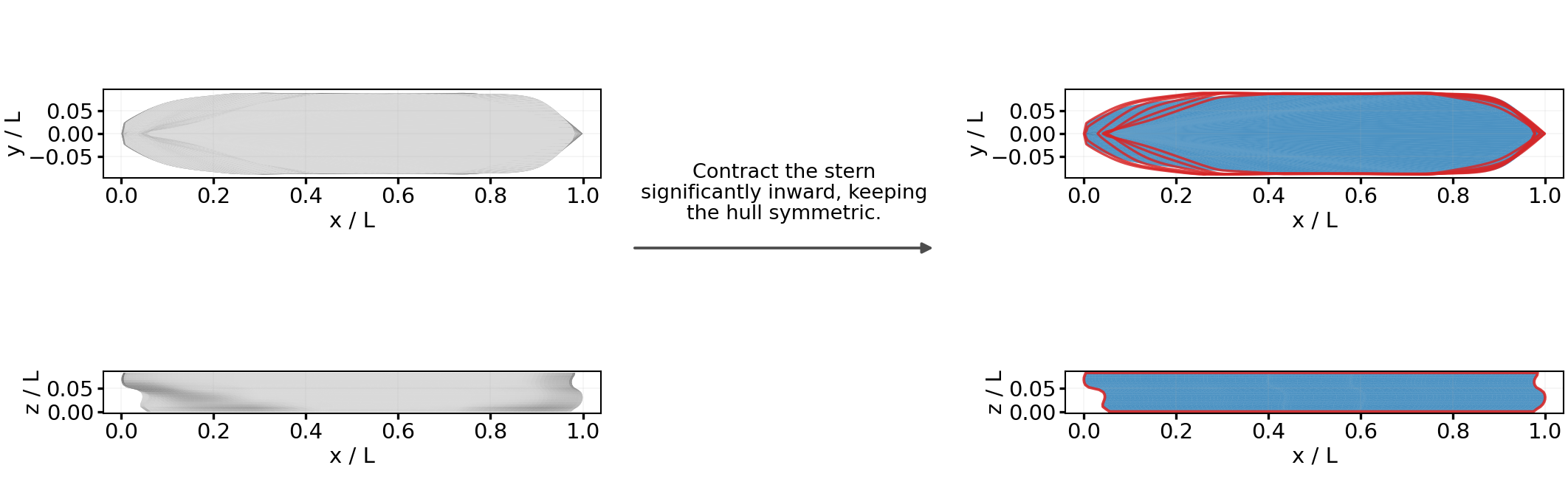}
        \caption{Single edit.}
    \end{subfigure}
    \par\smallskip
    \begin{subfigure}[t]{0.98\textwidth}
        \centering
        \includegraphics[width=\linewidth]{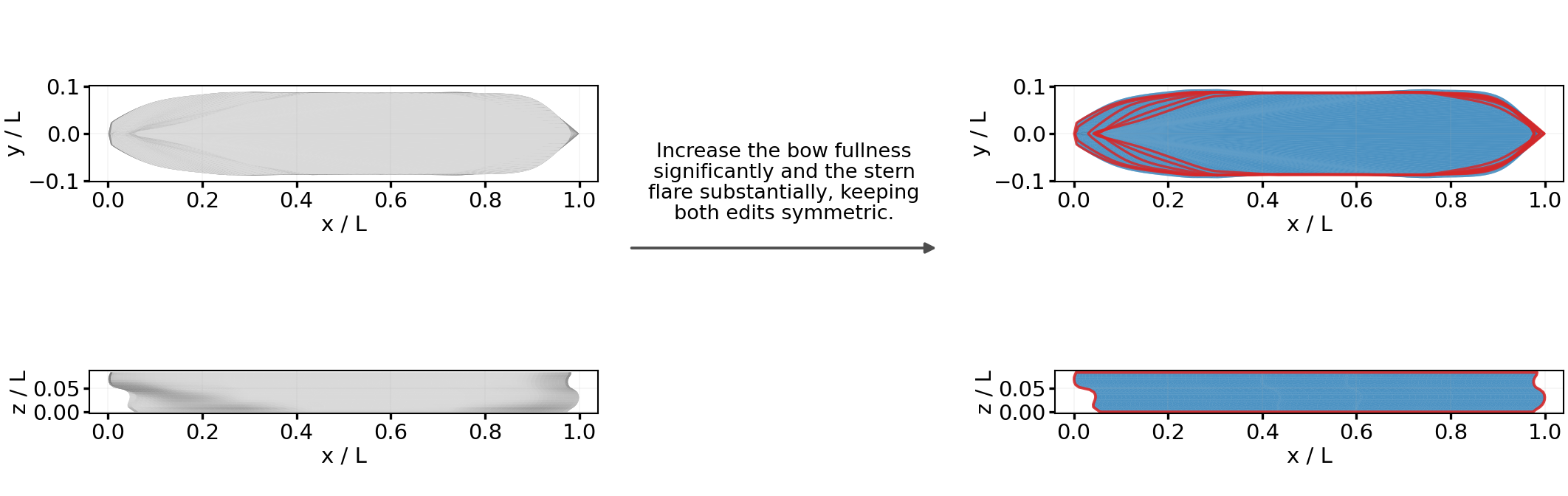}
        \caption{Multi-action edit.}
    \end{subfigure}
    \par\smallskip
    \begin{subfigure}[t]{0.98\textwidth}
        \centering
        \includegraphics[width=\linewidth]{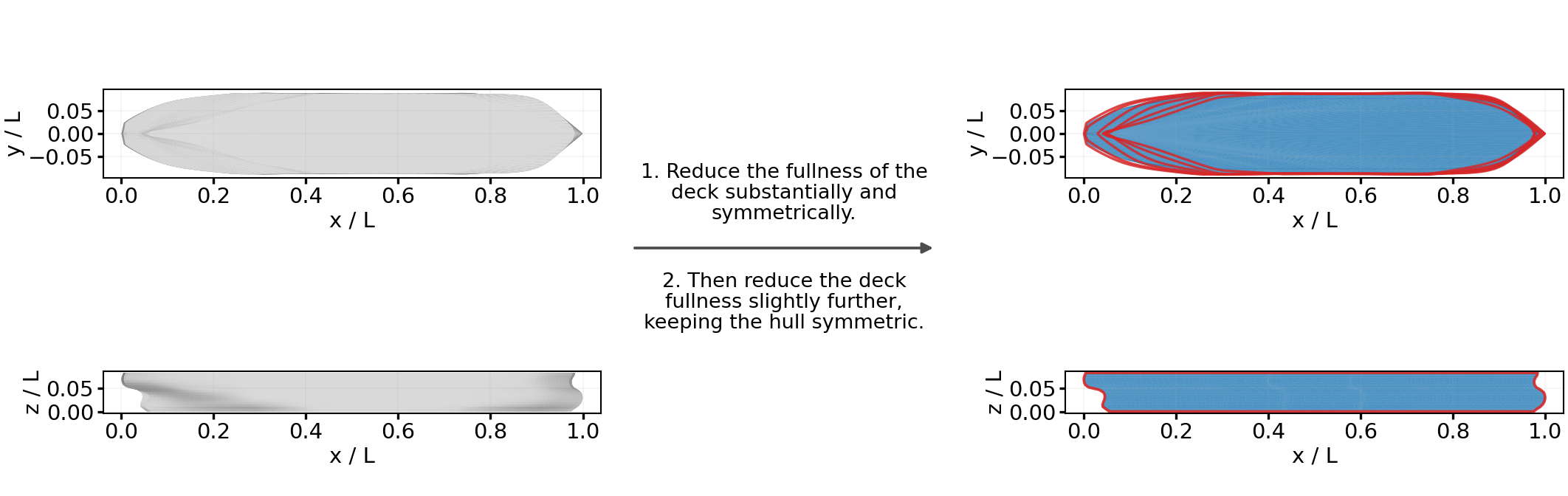}
        \caption{Continuous two-turn edit.}
    \end{subfigure}
    \caption{Alternative representative end-to-end editing cases from the same
    three interaction categories. The paired views show the original hull,
    natural-language instruction, and reconstructed modified hull. Original hulls
    are on the left and modified hulls (blue) are on the right; red lines trace
    sections of the original hull. Arrows indicate the editing direction.}
    \label{fig:end2end-alternative}
\end{figure*}

Figures~\ref{fig:end2end-primary} and~\ref{fig:end2end-alternative} show two
sets of representative edits. Each visualization places the natural-language
instruction between the original and modified hulls and presents both plan and
side views. The first set covers a single edit, a multi-action edit, and a
continuous two-turn edit. The second set provides alternative examples from the
same three interaction categories.

Table~\ref{tab:end2end-results} summarizes the three interaction settings.
All executed edits produced geometrically valid hulls, giving a 100\% geometry
validity rate in every category. Single edits achieved a 100\% end-to-end
success rate. Continuous two-turn edits achieved 98\%, with two constraint
violations among the 43 constrained experiments. Multi-region single-turn edits
achieved 60\% end-to-end success. Their execution and geometry checks reached
100\%, while 39 of 45 constrained experiments violated a constraint and one
experiment contained an action-sequence mismatch. This separation identifies the
constraint-handling stage as the main source of unsuccessful multi-region
trajectories in this evaluation.

\begin{table*}[t]
\caption{End-to-end editing results on the KVLCC2 hull. Magnitude-level accuracy
is calculated over all labelled action levels. Constraint satisfaction is
calculated over experiments containing at least one constraint.}
\label{tab:end2end-results}
\centering
\begin{tabular*}{\tblwidth}{@{\extracolsep{\fill}}lrrrrrr@{}}
\toprule
\shortstack{Interaction\\type} & Experiments & \shortstack{End-to-end\\success} & \shortstack{Trajectory\\exact} & \shortstack{Geometry\\valid} & \shortstack{Constraints\\satisfied} & \shortstack{Level\\accuracy} \\
\midrule
Single edit                 & 100 & 100.00\% & 100.00\% & 100.00\% & 100.00\% & 52.00\% \\
Multi-region, single turn  & 100 & 60.00\%  & 96.00\%  & 100.00\% & 13.33\%  & 70.48\% \\
Same-region, two turns     & 100 & 98.00\%  & 100.00\% & 100.00\% & 95.35\%  & 81.50\% \\
\bottomrule
\end{tabular*}
\end{table*}

The result demonstrates that the complete pipeline can transform natural-language
instructions into executable geometric edits and reconstruct valid hulls across
all three interaction settings. The distinction between trajectory matching and
end-to-end success is especially visible in multi-region editing, where the
individual execution remained valid while constraint satisfaction reduced the
number of accepted trajectories.

\subsubsection{Evaluation time}
\label{sec:end-to-end-latency}

We also evaluated the time required by each model to perform the ship design
decision task. Model and geometric processing were measured separately. For model processing,
we used the per-request latency recorded by each evaluator. The reported mean and
95th-percentile values include the model request and the evaluator-side response
handling; model initialization, result aggregation, and geometric processing are
excluded. The LLM evaluators recorded latency for both successful and failed
attempts.

\begin{table*}[t]
\caption{Per-request model latency on SDDBench. Mean and
95th-percentile latency are reported in milliseconds from the recorded evaluator
outputs. The API evaluators timed failed attempts as well as successful attempts;
the JEV latency statistics use its 4,322 completed requests.}
\label{tab:latency-results}
\centering
\begin{tabular*}{\tblwidth}{@{\extracolsep{\fill}}lrr@{}}
\toprule
Model & Mean latency (ms) & 95th percentile (ms) \\
\midrule
\multicolumn{3}{l}{\textit{Small language models}} \\
Qwen3                & 25280.05 & 46745.65 \\
Gemma3               & 18407.59 & 43119.55 \\
Llama3.2             & 22589.78 & 44708.37 \\
\hline
\multicolumn{3}{l}{\textit{Flagship language models}} \\
Claude Opus 5        & 7844.00  & 12531.86 \\
GPT-5.6 Sol          & 16009.20 & 36265.54 \\
DeepSeek V4.1 Flash  & 4630.63  & 11131.43 \\
\hline
\multicolumn{3}{l}{\textit{Language decision models}} \\
JEV                  & 1510.33  & 2908.01 \\
Chip (Ours)      & 118.01   & 235.55 \\
\bottomrule
\end{tabular*}
\end{table*}

Chip had the lowest recorded latency, with a mean of
118.01 ms and a 95th-percentile latency of 235.55 ms. JEV was the next fastest
language-decision model, while the flagship and small language models required
longer per-request processing under the evaluated settings. The geometric
processing stage added approximately 0.13 s per end-to-end experiment on
average: 0.114 s for a single edit, 0.123 s for a multi-region single-turn edit,
and 0.148 s for a same-region two-turn edit. These measurements show that Chip
generates ship design decisions at very low latency under the evaluated settings.

\subsection{Ablation Studies}
\label{sec:ablations}

To isolate the contributions of geometric representation, model scale, and
training-data size, we evaluated each design choice under the SDDBench protocol
and compared the resulting reconstruction and decision metrics.

\subsubsection{Sensitivity to NURBS Representation}
\label{sec:ablation-nurbs}

\begin{figure*}[t]
    \centering
    \begin{subfigure}[t]{0.98\textwidth}
        \centering
        \includegraphics[width=\linewidth]{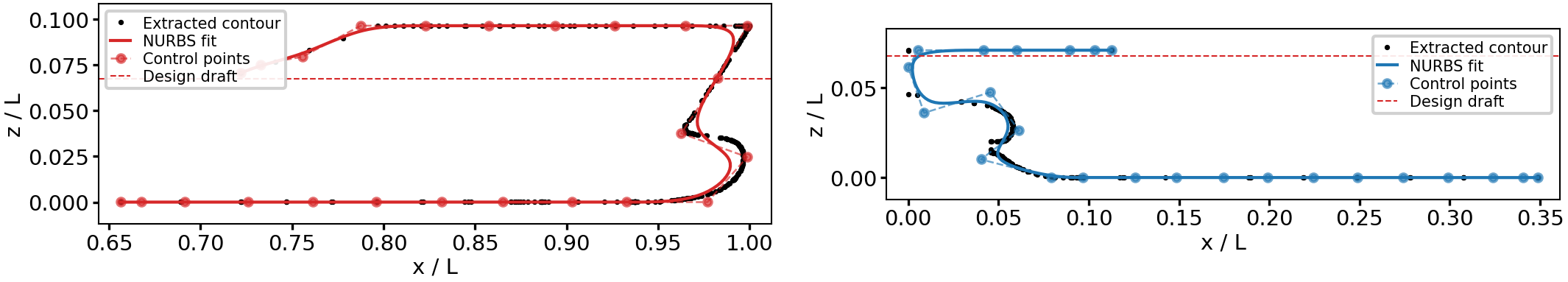}
        \caption{24 controls.}
    \end{subfigure}
    \par\smallskip
    \begin{subfigure}[t]{0.98\textwidth}
        \centering
        \includegraphics[width=\linewidth]{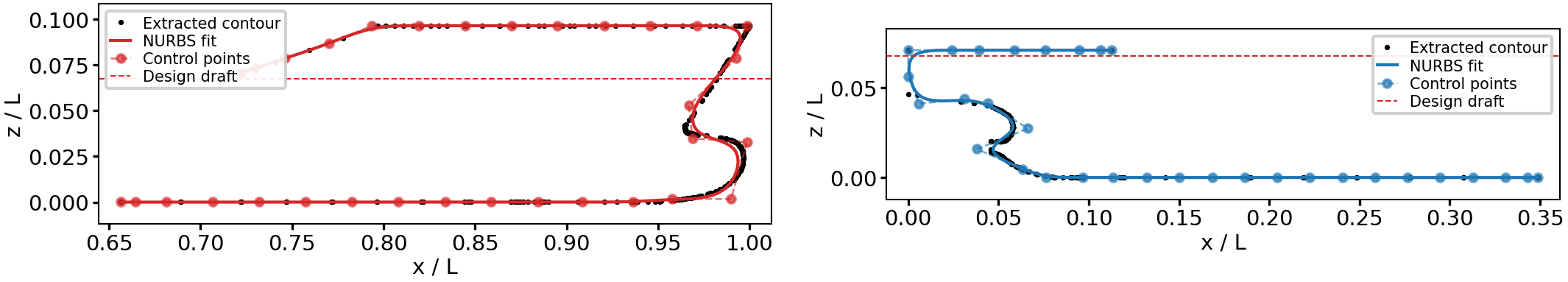}
        \caption{32 controls.}
    \end{subfigure}
    \par\smallskip
    \begin{subfigure}[t]{0.98\textwidth}
        \centering
        \includegraphics[width=\linewidth]{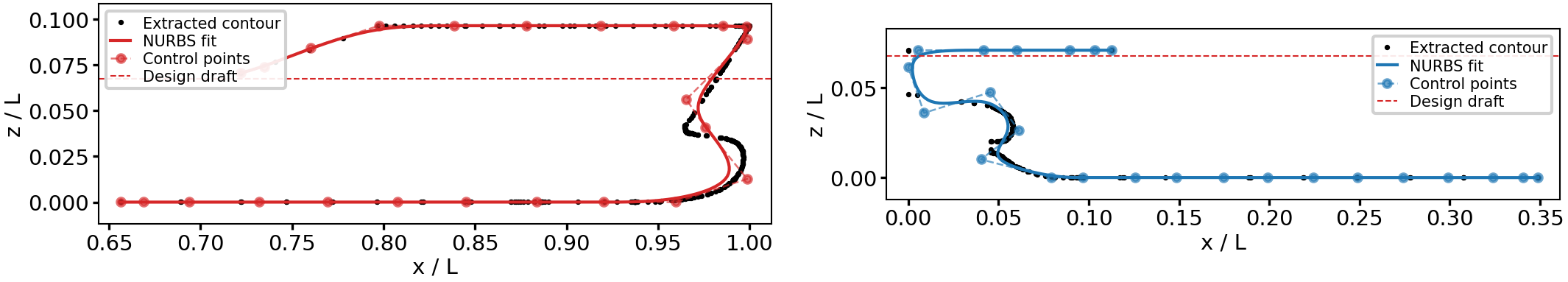}
        \caption{Deck knot.}
    \end{subfigure}
    \par\smallskip
    \begin{subfigure}[t]{0.98\textwidth}
        \centering
        \includegraphics[width=\linewidth]{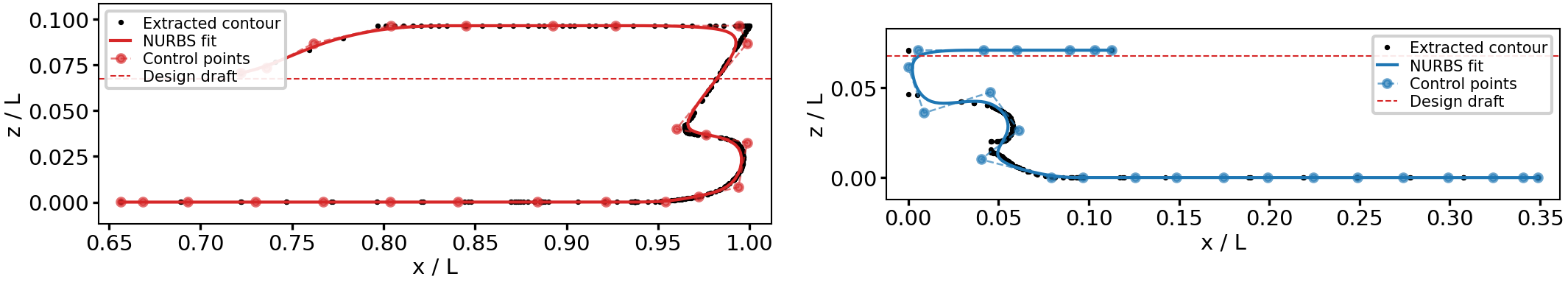}
        \caption{Bulb knots.}
    \end{subfigure}
    \par\smallskip
    \begin{subfigure}[t]{0.98\textwidth}
        \centering
        \includegraphics[width=\linewidth]{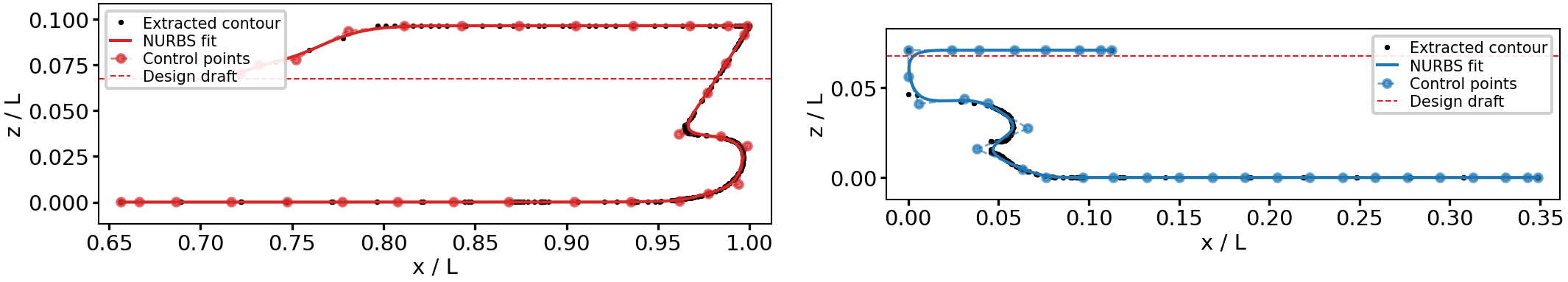}
        \caption{Combined.}
    \end{subfigure}
    \caption{KCS center-plane profile reconstruction for the five controlled
    representation variants. Each panel compares the extracted stem and stern
    contours with their fitted cubic-spline curves. Variants occupy separate rows,
    with stem contours on the left and stern contours on the right.}
    \label{fig:profile-ablation}
\end{figure*}

To isolate the effect of the center-plane profile representation on hull
reconstruction, we compared five configurations evaluated on the same 12 hulls,
with identical waterline levels, extracted profile samples, and mesh-comparison
settings. The 24-control configuration used 24 control points per stem and stern
contour with uniform clamped cubic-spline knots. Three representation operations
were then introduced separately or together, using the labels in
Figure~\ref{fig:profile-ablation}: ``32 controls'' increases both contours to 32
control points; ``Deck knot'' repeats the B-spline knot at the stem deck-corner
parameter, increasing its local multiplicity; and ``Bulb knots'' densifies the
interior stem-knot distribution around the bulb parameter. The ``Combined''
configuration applies all three operations.

The primary metrics were the RMSEs of the fitted stem and stern contours, calculated against their 80-point resampled reference contours. These
metrics directly measure the accuracy of the profile curves modified in this
experiment.

Table~\ref{tab:profile-ablation} reports both the KCS result and the arithmetic
mean over all 12 hulls. Increasing the control count reduced both stem and stern
errors, while the combined representation produced the lowest stem-contour RMSE
in both scopes: 0.000343 for KCS and 0.000221 averaged over the hull set. The
Deck knot and Bulb knots mainly affected the stem contour, with little change in
the stern-contour error. Figure~\ref{fig:profile-ablation} shows
the corresponding KCS reconstructions and the local changes around the bow
features.

\begin{table*}[t]
\caption{Ablation of the center-plane profile representation. The ``32 controls''
operation increases the stem and stern contours from 24 to 32 control points.
The ``Deck knot'' operation repeats the stem deck-corner knot, and ``Bulb knots''
densifies the interior stem-knot distribution around the bulb. ``Combined''
contains all three operations. Errors are normalized by the source-hull length;
``Mean'' is the arithmetic mean over the 12 classic hulls.}
\label{tab:profile-ablation}
\centering
\begin{tabular*}{\tblwidth}{@{\extracolsep{\fill}}lrrrrrrr@{}}
\toprule
Variant & \shortstack{32\\controls} & \shortstack{Deck\\knot} & \shortstack{Bulb\\knots} & \shortstack{KCS stem\\RMSE} & \shortstack{KCS stern\\RMSE} & \shortstack{Mean stem\\RMSE} & \shortstack{Mean stern\\RMSE} \\
\midrule
24 controls & -- & -- & -- & 0.002326 & 0.001581 & 0.001551 & 0.001236 \\
32 controls & \checkmark & -- & -- & 0.001222 & \textbf{0.000818} & 0.000909 & \textbf{0.000785} \\
Deck knot & -- & \checkmark & -- & 0.002187 & 0.001581 & 0.000994 & 0.001236 \\
Bulb knots & -- & -- & \checkmark & 0.001684 & 0.001581 & 0.001396 & 0.001236 \\
Combined & \checkmark & \checkmark & \checkmark & \textbf{0.000343} & \textbf{0.000818} & \textbf{0.000221} & \textbf{0.000785} \\
\bottomrule
\end{tabular*}
\end{table*}

\subsubsection{Effect of Chip Model Scale}
\label{sec:ablation-scale}

We next evaluated the effect of model scale while keeping the data, probability
questions, inference protocol, action projection, and SDDBench fixed. The
comparison used Chip models with 0.8 billion, 2 billion, and 4 billion parameters.
The question accuracy was computed over the successfully scored probability
questions, and FFD exact match was computed over all requested turns, including
failed evaluations. NLL, ECE, and Brier score measure probability quality and
calibration.

Table~\ref{tab:scale-ablation} shows closely clustered results across the three
models. Chip-2B achieved the highest question accuracy and the lowest probability
error measures, whereas Chip achieved the highest FFD exact match. Because
ship design is formulated as a constrained decision problem, its decision
freedom is relatively limited; consequently, the evaluated model sizes did not
show marked performance differences.

\begin{table*}[t]
\caption{Chip model-scale ablation on SDDBench. Question accuracy uses the
successfully scored questions. FFD exact match uses all requested turns. Best values in each column are bold.}
\label{tab:scale-ablation}
\centering
\begin{tabular*}{\tblwidth}{@{\extracolsep{\fill}}lrrrrrr@{}}
\toprule
Model & Completed records & Question accuracy & FFD exact match & NLL & ECE & Brier score \\
\midrule
Chip & 5,000 & 95.8962\% & \textbf{99.32\%} & 0.0951 & 0.0032 & 0.0551 \\
Chip-2B   & 5,000 & \textbf{95.9192\%} & 99.30\% & \textbf{0.0948} & \textbf{0.0022} & \textbf{0.0548} \\
Chip-4B   & 4,999 & 95.8794\% & 99.24\% & 0.0951 & 0.0025 & 0.0549 \\
\bottomrule
\end{tabular*}
\end{table*}

\subsubsection{Effect of Training Data Size}
\label{sec:ablation-data}

To assess the contribution of the training-set size, we held the Chip
architecture fixed and varied the amount of ship-design training data. Nested
subsets containing 1\%, 2\%, 4\%, 8\%, and 25\% of the SDD Dataset training
partition were sampled using a fixed random seed. Each smaller subset was contained in every larger subset. The training
configuration was identical to the main Chip experiment except for the amount of
training data. With the epoch count fixed, the number of optimizer updates
increased from 222 at 1\% to 22,152 at 100\%. Kev-Base supplied the 0\% reference
before ship-design adaptation.

All seven checkpoints were evaluated on SDDBench using the probability-question
and FFD action-scoring protocol described in
Section~\ref{sec:ship-decision-results}. Table~\ref{tab:data-scaling} combines
the subset evaluations with the full-data result reported in
Section~\ref{sec:ablation-scale}. Figure~\ref{fig:data-scaling} plots the two
accuracy measures against the same training-data shares. The question accuracy
uses successfully scored questions, while FFD exact match includes all SDDBench
records.

\begin{table*}[t]
\caption{Effect of SDD Dataset size on Chip. Training share is relative to the
SDD Dataset training partition; the 0\% row evaluates Kev-Base before ship-design
adaptation. Accuracy values are
percentages. Lower values are preferred for NLL, ECE, and Brier score. Best
values are shown in bold.}
\label{tab:data-scaling}
\centering
\begin{tabular*}{\tblwidth}{@{\extracolsep{\fill}}rrrrrrr@{}}
\toprule
\shortstack{Training\\share (\%)} & \shortstack{Training\\records} &
\shortstack{Question\\accuracy $\uparrow$} & FFD exact match $\uparrow$ &
NLL $\downarrow$ & ECE $\downarrow$ & Brier $\downarrow$ \\
\midrule
0   & 0      & 80.82 & 72.24 & 0.6972 & 0.1995 & 0.3439 \\
1   & 886    & 94.02 & 96.86 & 0.1729 & 0.0179 & 0.0850 \\
2   & 1,772  & 94.57 & 97.88 & 0.1533 & 0.0110 & 0.0754 \\
4   & 3,544  & 95.13 & 98.64 & 0.1331 & 0.0105 & 0.0682 \\
8   & 7,088  & 95.39 & 98.82 & 0.1196 & 0.0097 & 0.0642 \\
25  & 22,151 & 95.66 & 99.14 & 0.1040 & 0.0047 & 0.0584 \\
100 & 88,604 & \textbf{95.90} & \textbf{99.32} & \textbf{0.0951} &
\textbf{0.0032} & \textbf{0.0551} \\
\bottomrule
\end{tabular*}
\end{table*}

\begin{center}
    \includegraphics[width=\linewidth]{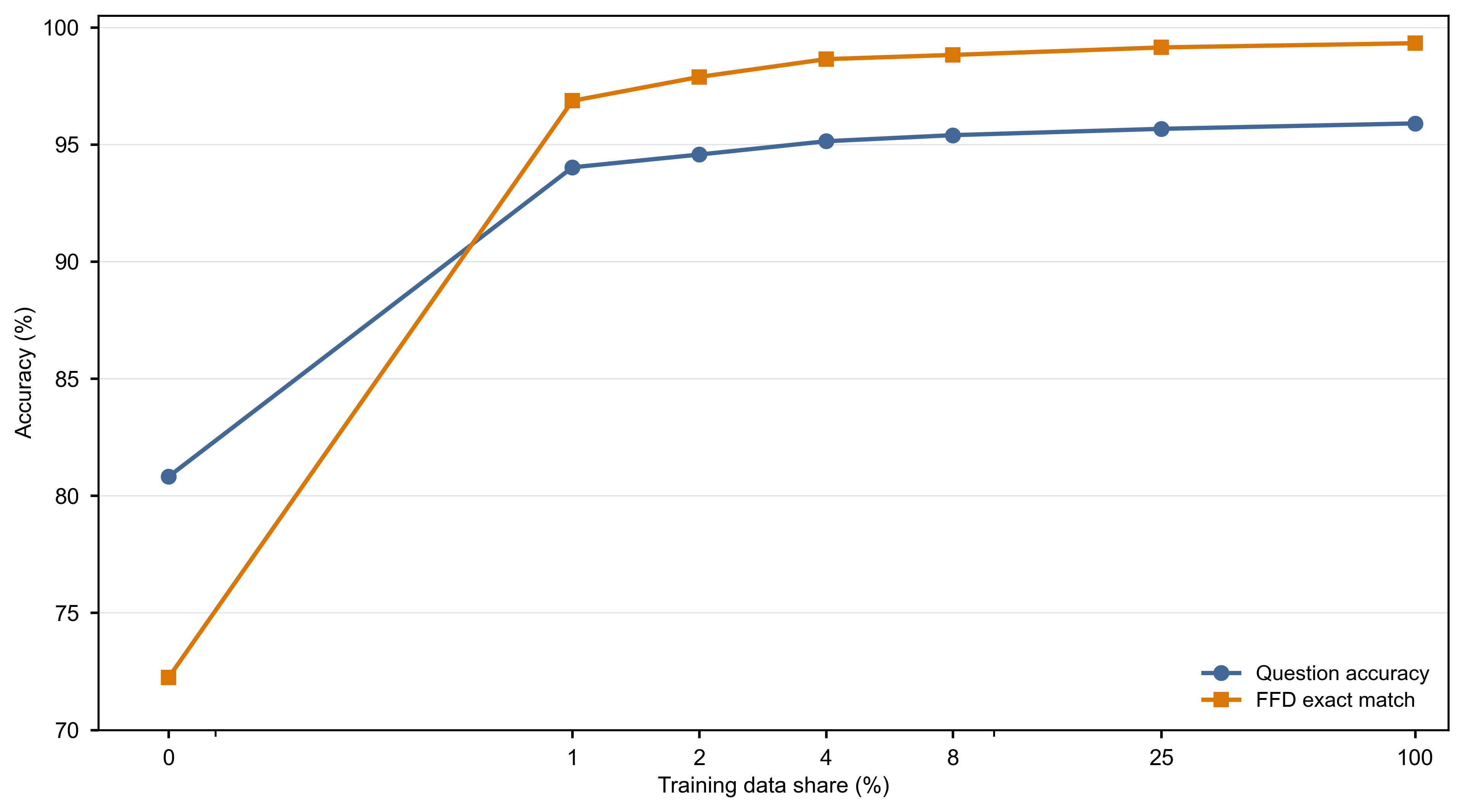}
    \captionof{figure}{Effect of training-data size on Chip accuracy. The
    horizontal axis shows the share of the SDD Dataset training partition; the
    0\% point is the Kev-Base reference before ship-design adaptation.}
    \label{fig:data-scaling}
\end{center}

Ship-design adaptation produced a large initial improvement: the 1\% model
reached 94.02\% question accuracy and 96.86\% FFD exact match, compared with
80.82\% and 72.24\% for the 0\% reference. Across the trained checkpoints from
1\% to 100\%, both accuracies increased, while NLL, ECE, and Brier score
decreased. The full-data model attained the best values in the table.

The gains became smaller at larger data sizes, yet the full training set still
produced the best measured performance. From 1\% to 100\% of the original
training dataset, question accuracy increased by 1.88 percentage points and FFD
exact match increased by 2.46 percentage points, while NLL, ECE, and Brier score
all decreased. Together with Section~\ref{sec:ablation-scale}, these results
show that, for the ship-design decision task, expanding the training dataset
produced larger measured gains than increasing model size across the tested
range.

\section{Discussion}
\label{sec:discussion}

\subsection{Constraint Semantics in End-to-End Ship-Form Editing}
\label{sec:end2end-discussion}
The end-to-end experiments exposed a failure mode in the interaction between
action composition and constraint enforcement. Multi-region single-turn editing
achieved a 60\% end-to-end success rate even though execution and geometry checks
both reached 100\%. Among the 45 experiments containing at least one constraint,
39 violated a constraint. Same-region two-turn editing reached 98\% success, with
two additional constraint violations. The concentration of failures in the
constraint stage points to the semantics of sequential constraint handling as a
central factor in the end-to-end result.

The implementation applies actions sequentially and performs displacement
compensation inside each constrained action. Before such an action, the executor
records the current volume; after the deformation, it rescales the hull to that
recorded value. The end-to-end checker then evaluates the resulting turn against
the volume at the beginning of the turn whenever any action in that turn carries
a displacement-preservation requirement. Consequently, action-level compensation
and turn-level verification use different reference states. An unconstrained
action in the same turn can change the volume after a compensated action, and the
final turn can therefore fail the global displacement check despite correct
individual action semantics. The deck-line check follows the same sequential
state logic and uses a strict numerical tolerance on the final top controls.

This mechanism is supported by the failure records. Across the 41 constraint
violations, 34 cases had both the predicted action semantics and the predicted
constraint fields matched to the reference actions. The violations comprised 31
volume-only cases, six deck-line-only cases, and four cases containing both types
of residual. In the multi-region single-turn subset, the action list contained
two or three sequential edits, allowing unconstrained edits to alter a quantity
that an earlier constrained edit had restored. These cases show that trajectory
matching and geometric validity can coexist with failure of the current
constraint contract. A turn-level constraint representation, or a final
projection that enforces all active constraints after the complete action list,
would align execution with the end-to-end criterion.

\subsection{Why Large Language Models Fail in Ship Free-Form Deformation Decision Making}
\label{sec:llm-failures}
SDDBench places two coupled requirements on a language model: it must infer
the ship-form action and serialize a complete typed probability response. The
evaluator first receives the model response, parses JSON, checks the required
probability keys and value ranges, verifies that each
distribution sums to one, and then projects the valid response into an ordered
action sequence. A response can therefore fail after generation during parsing,
probability-schema validation, or action projection.

The local small models failed predominantly at these post-generation interface
stages. Of 5,000 requests, Qwen3 had 4,225 rejected records, Gemma3 had 2,359,
and Llama3.2 had 4,443. For Qwen3, the largest rejection groups were mismatched
keys for the region, action-count, and operation distributions, with 2,463,
977, and 772 cases, respectively. Gemma3 produced 1,419 action-count key
mismatches, 694 operation key mismatches, and 230 region key mismatches. Llama3.2
produced 1,820 action-count key mismatches, 1,388 responses without the required
answer object, 696 operation key mismatches, and 259 region key mismatches.
These records establish that output serialization is a major source of failure
for the small local models.

Valid responses still contained decision errors. The question accuracy was 75.61\%
for Qwen3, 66.97\% for Gemma3, and 63.94\% for Llama3.2, while FFD exact match scores
were 2.84\%, 15.98\%, and 2.16\%, respectively. The action metric uses all
requested turns, so it incorporates both rejected responses and incorrect
actions; the question metric isolates decisions that passed the probability
checks. The two metrics together show separate costs for semantic action
selection and strict response construction. The latency results add a system
cost: the mean per-request latencies of these models ranged from 18.41 to
25.28~s, while the Chip model required 118.01~ms under the same SDDBench setting. A decision interface for ship-form editing therefore benefits
from training that jointly covers domain semantics, action ordering, constraint
labels, and the required probability schema.

\subsection{Specialized Model Scale and Data Efficiency}
\label{sec:scale-efficiency}
The scale and data ablations separate capacity from task-specific coverage. With
the same SDD Dataset and evaluation protocol, increasing Chip from 0.8 billion
to 2 billion and 4 billion parameters changed question accuracy by only 0.0398
percentage points across the three models. Chip-2B gave the highest question
accuracy and the best NLL, ECE, and Brier score, whereas Chip gave the highest FFD exact match. The close results
show that the tested decision space can be represented by Chip, and that additional parameters alone produced little movement in the
measured metrics.

Training-data size had a stronger measured effect. Ship-design adaptation raised
question accuracy from 80.82\% at the 0\% reference to 94.02\% with 1\% of the
original training dataset, followed by continued improvement through the full
dataset. From 25\% to 100\%, question accuracy increased from 95.66\% to 95.90\%
and FFD exact match increased from 99.14\% to 99.32\%; NLL, ECE, and Brier
score also decreased. The continued change
at the largest evaluated data size indicates that the training examples supplied
useful coverage for both action decisions and probability calibration.

The task has a finite action vocabulary but a structured composition: each
request specifies an action count, regions, operations, qualitative magnitude
levels, and optional preservation constraints, and the resulting sequence must
remain compatible with the free-form deformation executor. This structure makes
capacity sufficient for the tested action space while placing high demands on
label coverage, action ordering, and output regularity. The ablations therefore
support a data--capacity trade-off in which task-specific data and a compatible
output protocol provide larger gains than parameter growth across the tested
range. Chip combines this performance
with the lowest measured inference latency, making it a practical capacity point
for the evaluated decision interface.

\section{Conclusion and Future Work}
\label{sec:conclusion}

This work presents NL2Hull Framework as a discrete constrained decision system
and implements the complete path from natural-language requests to typed action
probabilities, executable FFD, and reconstructed hull checks through the CFFD
Engine. The system combines a NURBS waterline representation with an explicit
action projection and the SDDBench evaluation protocol. On SDDBench, Chip achieved 95.90\% question accuracy and 99.32\% FFD exact
match with well-calibrated probabilities. The representation study reduced the mean
stem-contour error to 0.000221 over 12 hulls, and the end-to-end study verified
that the system can execute single edits and multi-turn edits on a KVLCC2 hull.

The end-to-end experiments also identified a concrete CFFD Engine limitation.
Displacement preservation is compensated at the action level, while the final
turn is checked against the state at the beginning of the turn. Unconstrained
actions composed with constrained actions can therefore alter the final volume,
and strict deck-line checks can retain small numerical residuals after otherwise
valid deformation. The CFFD Engine should be extended with composition-aware
constraint handling, final-state projection, and tolerances linked to the
reconstruction and meshing resolution. These changes would improve the
flexibility and generality of sequential edits across hull forms and constraint
combinations.

The current decision interface also represents qualitative magnitude through an
ordinal level and uses default regional extents; it does not predict an explicit
continuous deformation magnitude or spatial extent. Future model development
should add continuous numerical outputs, train them against geometric outcomes,
and evaluate their calibration together with discrete action accuracy. A more
flexible CFFD Engine and a continuous-control decision model would extend
NL2Hull Framework from validated typed edits toward broader ship-form design
workflows.

\clearpage
\onecolumn
\appendix

\section{Prompt templates}
\label{app:prompts}

The prompts actually used in the code are written in Chinese. We translate them
into English here to improve readability; the translations below document the
runtime prompts without changing them. The prompts are displayed in verbatim code
blocks, which provide the LaTeX equivalent of fenced Markdown code blocks. Braced
fields are replaced with the current request, typed questions, or structured
actions.

\subsection{SDDBench probability-output prompt}
\label{app:benchmark-prompt}

Each question block supplies its identifier, type, instruction, and all candidate
keys with their descriptions. The block format is
\begin{verbatim}
Question ID: {question identifier}
Type: {choice, noul, or score}
Instruction: {question instruction}
Options (keys must be preserved exactly):
- {candidate key}: {candidate description}
\end{verbatim}
The blocks are inserted into the following translated rendering of the Chinese
runtime prompt:

\begingroup
\small
\begin{verbatim}
You are a ship design decision model. Based on the state and questions,
output a probability distribution for every question.

State:
{record state}

Questions:
{one question block per typed question}

Output strict JSON only. Do not return Markdown, explanations, or extra text:
{"answers": {
  "question_id": {"probabilities": {"option_key": 0.5}}
}}

Requirements:
- Answer every question; question IDs must match exactly.
- The keys in every probabilities object must match the keys listed for that question exactly.
- Every probability must be a number between 0 and 1, and probabilities for each question must sum to 1.
- For choice questions, use option names as keys; for noul questions, use false and true;
  for score questions, use strings such as 0, 1, and 2 as keys.
- Return only the JSON object above.
\end{verbatim}
\endgroup

\subsection{Natural-language description-generation prompt}
\label{app:description-prompt}

The template is applied once per editing turn and language variant. The structured
action field contains every action requested in the current turn, in execution
order. The translated prompt is
\begingroup
\small
\begin{verbatim}
You are generating one Chinese natural-language user request for a ship-form
deformation dataset.
{previous-turn context, if present}
{single-action or multi-action instruction}
{background instruction}

This is the {variant number} natural expression of the same structured action.
Use a sentence structure different from the other expressions while preserving
accuracy. You may vary sentence structure, verbs, degree adverbs, and connectors.
Do not translate one change level into one fixed word; all expressions must retain
the same relative magnitude.

Requirements:
1. Return one natural and concise Chinese user sentence only. Do not return JSON,
   explanations, a title, bullet points, or quotation marks.
2. Preserve the action region, operation, relative deformation magnitude, and
   constraint meaning. Do not add engineering objectives absent from the input JSON.
3. For actions with an internal intensity code, use natural qualitative wording.
   The output must not contain internal codes, numeric intensity, magnitude, level,
   grade, intensity, or tier terminology. The wording must preserve the relative strength.
4. If magnitude_level is null, reproduce every value in magnitude_value,
   longitudinal_extent, and vertical_extent character by character. Do not round,
   truncate, or rewrite decimals. In this case, words such as numerical change may be used.
5. Express constraints such as preserve_displacement and preserve_deck_line in the sentence.
6. Do not introduce physical units such as metres or millimetres; the data use normalized parameters.
7. Use the fixed regional vocabulary: bow, stern, bulb, midbody, deck, bilge, and global.
8. Use the fixed directional vocabulary: forward, aftward, upward, downward, outward, and inward.
9. Use the fixed operation vocabulary for fullness, flare, length, and breadth changes.
10. change_bulb_length means changing bulb length by the specified value; magnitude_value
    is the change, not the final length.

Hull: {hull identifier}
Interaction type: {interaction type}
Current action JSON:
{structured actions}
\end{verbatim}
\endgroup

For a single action, the runtime instruction is ``Please express only the current
action.'' For multiple actions, it is ``One input contains changes to multiple
regions; express all actions completely in one sentence.'' A later turn includes
this context and the previous actions:
\begin{verbatim}
This is a follow-up edit in the same region. The previous action is shown below.
The current description should express only this turn; a connector such as "again"
may be used:
{previous structured actions}
\end{verbatim}

The background instruction either requests a direct edit without additional
background or selects one of the following scenario openings while prohibiting new
design goals:
\begin{verbatim}
I am designing a ship using {hull identifier} as the parent hull, and I need to ...
You are a ship-design expert. Please ...
I am conducting AI4CFD research and need to generate ship-form data. Please ...
For optimization of the {hull identifier} parent hull, please ...
\end{verbatim}

If validation fails, the following instruction is appended to the original prompt:
\begin{verbatim}
The previous answer failed format or semantic validation. Generate again and return
only one Chinese user sentence satisfying all requirements. Do not use the words
magnitude, level, grade, intensity, or tier, and do not explain the validation process.
\end{verbatim}

\clearpage
\section{Chip training configuration}
\label{app:training}

Chip was initialized from the open-sourced Kev-0.8B decision checkpoint, whose
language backbone is Qwen3.5-0.8B-Base. Ship-design adaptation used all 88,604
records in the SDD Dataset training partition, 2 epochs, rank-16 low-rank
adaptation, and a learning rate of $2\times10^{-5}$. A microbatch of two records
with four gradient-accumulation steps gave an effective batch size of eight.
Weights and computation used 16-bit brain floating point (bfloat16), with gradient
checkpointing enabled, on an NVIDIA RTX A6000 graphics processor.

Table~\ref{tab:chip-training-config} lists the language-backbone and batch
settings for the model-scale experiment. Chip-2B used a decision checkpoint trained on the
original Kev decision dataset; this preliminary training used two epochs, a
learning rate of $5\times10^{-5}$, rank-16 low-rank adaptation, and the same
effective batch size of eight. Chip-4B used the open-sourced Kev-4B decision checkpoint.
All three variants then used the same SDD Dataset training partition and
ship-design adaptation settings described above, with the batch and accumulation
settings shown in the table.

\begin{table*}[t]
\caption{Variant-specific language-backbone and batch settings for ship-design
adaptation. The batch size is the number of records per microbatch; accumulation
is the number of gradient-accumulation steps.}
\label{tab:chip-training-config}
\centering
\begin{tabular*}{\tblwidth}{@{\extracolsep{\fill}}lp{0.45\textwidth}rr@{}}
\toprule
Model & Language backbone & Batch size & Accumulation \\
\midrule
Chip & Qwen3.5-0.8B-Base & 2 & 4 \\
Chip-2B & Qwen3.5-2B-Base & 2 & 4 \\
Chip-4B & Qwen3.5-4B-Base & 1 & 8 \\
\bottomrule
\end{tabular*}
\end{table*}

For the training-data-size experiment, every adapted checkpoint was independently
initialized from the same Kev-0.8B checkpoint and used the main Chip
training settings. Nested subsets were sampled with random seed 0 at training
shares of 1\%, 2\%, 4\%, 8\%, 25\%, and 100\%, corresponding to 886, 1,772,
3,544, 7,088, 22,151, and 88,604 records.

\bibliographystyle{cas-model2-names}
\bibliography{references}

\end{document}